\pdfoutput=1

\documentclass[11pt]{article}
\PassOptionsToPackage{table}{xcolor}

\usepackage[preprint]{acl}
\usepackage[table]{xcolor}
\usepackage{amssymb}
\usepackage{array}
\usepackage{times}
\usepackage{latexsym}
\usepackage{xspace}
\usepackage{dsfont} 
\usepackage[T1]{fontenc}
\usepackage[utf8]{inputenc}

\usepackage[many]{tcolorbox}

\usepackage[utf8]{inputenc}

\usepackage{microtype}
\usepackage{microtype}
\usepackage[export]{adjustbox}

\usepackage{inconsolata}

\usepackage{graphicx}

\usepackage{tikz}
\usepackage{natbib}
\usepackage{float}
\usepackage{mathtools}
\usepackage{booktabs}
\usepackage{pifont}
\definecolor{darkred}{rgb}{0.8, 0, 0}  
\definecolor{darkgreen}{rgb}{0, 0.5, 0} 
\definecolor{red}{rgb}{1, 0, 0}
\usepackage{multirow}
\usepackage{makecell}
\usepackage{enumitem}
\usepackage{cleveref}
\usepackage{xspace}
\usepackage{graphicx}

\newcommand{\redmark}[1]{%
  \begingroup\setlength{\fboxsep}{0.5pt}%
  \colorbox{red!30}{\vphantom{Ay}\vphantom{Ay}#1}%
  \endgroup
}
\newcommand{\greenmark}[1]{%
  \begingroup\setlength{\fboxsep}{0.5pt}%
  \colorbox{darkgreen!30}{\vphantom{Ay}\vphantom{Ay}#1}%
  \endgroup
}

\newcommand{\caparena}{\textit{CapArena}}
\newcommand{\caparenaA}{\textit{CapArena-Auto}}
\usepackage{fdsymbol}
\usepackage{fontawesome}
\newcommand{\fstar}{\textsuperscript{\fontsize{6pt}{6pt}\selectfont \faMoonO}}

\title{CapArena: Benchmarking and Analyzing Detailed Image Captioning \\ in the LLM Era}

\author{
Kanzhi Cheng\textsuperscript{$\diamondsuit$} \footnotemark[2] \quad
Wenpo Song\textsuperscript{$\diamondsuit$} \footnotemark[2] \quad 
Jiaxin Fan\textsuperscript{$\diamondsuit$} \footnotemark[2] \quad 
Zheng Ma\textsuperscript{$\diamondsuit$} \quad
Qiushi Sun\fstar~ \\
\bf{
Fangzhi Xu\textsuperscript{$\heartsuit$} \quad
Chenyang Yan\textsuperscript{$\diamondsuit$} \quad
Nuo Chen\textsuperscript{$\diamondsuit$} \quad
Jianbing Zhang\textsuperscript{$\diamondsuit$} \quad
Jiajun Chen\textsuperscript{$\diamondsuit$}
}\\
\textsuperscript{$\diamondsuit$}National Key Laboratory for Novel Software Technology, Nanjing University \\
\fstar The University of Hong Kong 
\textsuperscript{$\heartsuit$}Shanghai Artificial Intelligence Laboratory \\
\texttt{chengkz@smail.nju.edu.cn} \quad \quad
\texttt{zjb@nju.edu.cn}
}

\begin{document}
\maketitle
\footnotetext[2]{Equal contribution.}
\begin{abstract}
Image captioning has been a longstanding challenge in vision-language research. 
With the rise of LLMs, modern Vision-Language Models (VLMs) generate detailed and comprehensive image descriptions.
However, benchmarking the quality of such captions remains unresolved.
This paper addresses two key questions: 
(1) How well do current VLMs actually perform on image captioning, particularly compared to humans?
We built \caparena, a platform with over 6000 pairwise caption battles and high-quality human preference votes. 
Our arena-style evaluation marks a milestone, showing that leading models like GPT-4o achieve or even surpass human performance, while most open-source models lag behind.
(2) Can automated metrics reliably assess detailed caption quality?
Using human annotations from \caparena, we evaluate traditional and recent captioning metrics, as well as VLM-as-a-Judge. 
Our analysis reveals that while some metrics (e.g., METEOR) show decent caption-level agreement with humans, their systematic biases lead to inconsistencies in model ranking.
In contrast, VLM-as-a-Judge demonstrates robust discernment at both the caption and model levels. 
Building on these insights, we release \caparenaA, an accurate and efficient automated benchmark for detailed captioning, achieving 94.3\% correlation with human rankings at just \$4 per test. 
Data and resources will be open-sourced at \href{https://caparena.github.io}{CapArena}.
\end{abstract}

\section{Introduction}

Image captioning, the task of generating textual descriptions for images, has long been a fundamental challenge in both the computer vision and natural language processing communities \cite{vinyals2015show, anderson2018bottom, cornia2020meshed}.
It has broad and valuable applications, such as assisting visually impaired individuals and supporting multimedia retrieval.
Driven by progress in LLMs, modern Vision-Language Models (VLMs) are capable of generating long, 
\begin{figure}[t]
  \centering
  \includegraphics[width=\columnwidth]{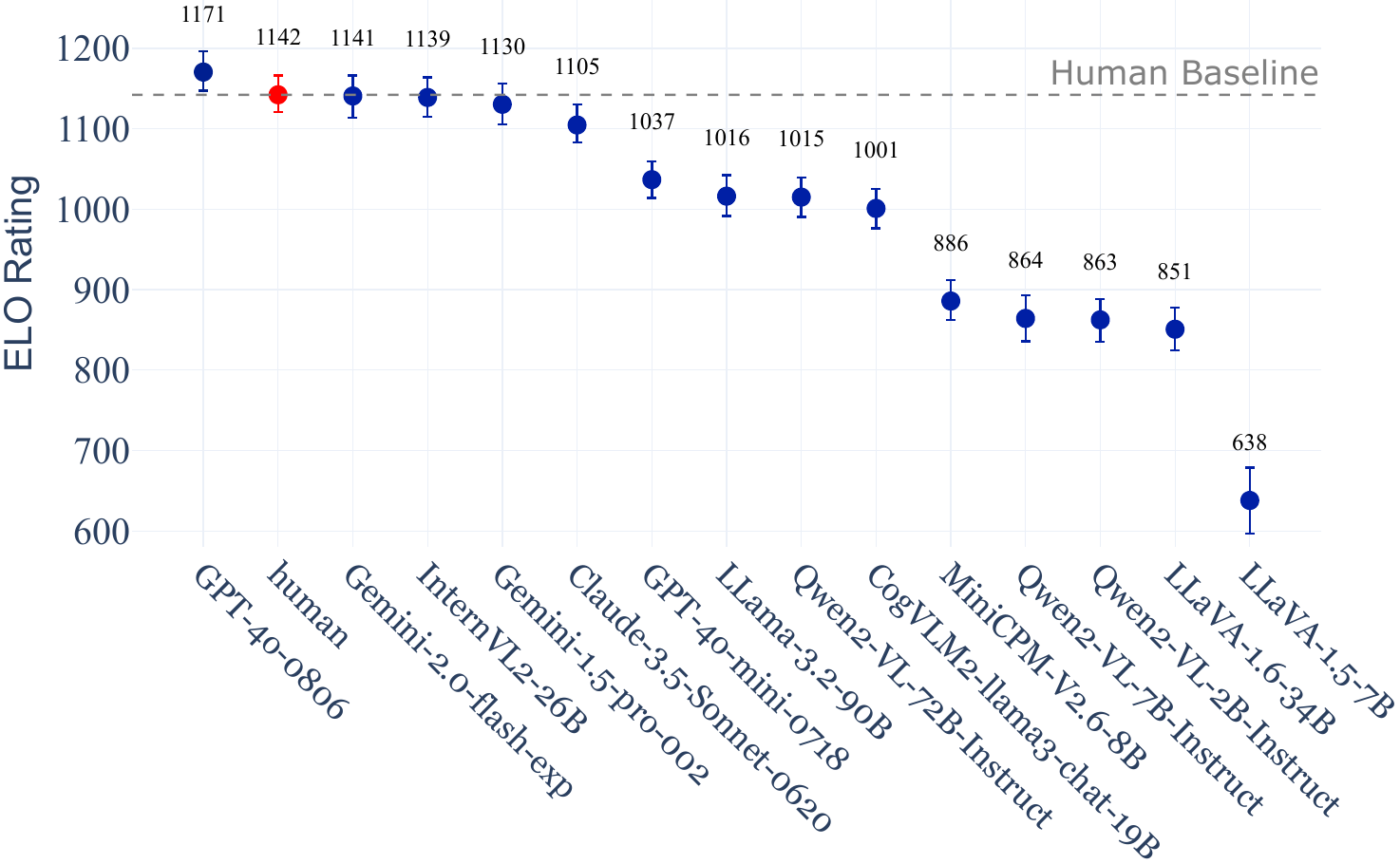} 
  \caption{Model rankings from \caparena{} in detailed captioning. Top models are comparable to humans, while most open-source models lag behind.}
  \label{fig:ranking}
\end{figure}
detailed descriptions of image content \cite{openai2023gpt4, chen2024far}, moving beyond the short captions of traditional methods.
The proliferation of VLMs presents new opportunities for the image captioning field.

However, image captioning has not advanced as expected.
Current VLMs focus on tasks like Visual Question Answering \cite{yue2024mmmu} and multimodal reasoning \cite{cheng2024vision}, bypassing the essential task of image captioning. 
A few works still rely on MSCOCO \cite{lin2014microsoft}—a dataset with an average caption length of 10 words, which is clearly outdated for evaluating advanced VLMs.
This obstacle stems from the inherent difficulty in evaluating detailed captions. Unlike multiple-choice questions or mathematical reasoning, captioning lacks explicit answers, resulting in the absence of reliable evaluation benchmarks.
Researchers are unable to assess the captioning capabilities of existing VLMs, nor effectively evaluate and improve their own models.

This paper addresses two key questions to drive the evolution of image captioning in the LLM era:
\emph{(1) How do existing VLMs perform in detailed captioning? Do top models achieve human-level performance, and how do they compare against each other?}
We conduct the first large-scale human evaluation to benchmark current VLMs.
Further, using resulting annotations from (1) as ground truth, we analyze existing captioning metrics and ask:
\emph{(2) How can we develop automated evaluation methods that reliably measure detailed caption quality and align with human preferences?}

To explore the first question, we developed \mbox{\caparena}, which includes over 6000 high-quality human annotations to evaluate the detailed captioning capabilities of 14 advanced VLMs and humans.
In our preliminary study, we found that traditional scoring methods are unsuitable for annotating detailed captions' quality due to their fine-grained intricacy and diversity.
Inspired by LLM evaluation \cite{chiang2024chatbot}, we adopted a pairwise caption battles paradigm.
The resulting model rankings are shown in \Cref{fig:ranking}. For the first time, we observe that state-of-the-art models, such as GPT-4o, are comparable to or even surpass human-level performance, marking a pivotal milestone in image captioning.
While open-source VLMs achieve competitive results on general benchmarks, CapArena reveals a persistent performance gap between them and commercial models.
An exception is InternVL2-26B \cite{chen2024far}, a smaller open-source model that stands out with its strong performance, underscoring the potential of compact and efficient VLMs for detailed captioning.

In response to the second question, we conducted a comprehensive analysis of traditional and recent captioning metrics, as well as the ability of VLM-as-a-Judge \cite{chen2024mllm} to assess caption quality.
We compared these metrics against human preferences from \caparena.
Our results reveal that most metrics designed for short captions, such as CLIPScore \cite{hessel2021clipscore}, fail entirely in the detailed captioning task.
Although some rule-based metrics, such as METEOR \cite{banerjee2005meteor}, exhibit decent agreement with human judgments at the caption level, they suffer from systematic biases across different VLMs.
This results in low agreement at the model level, where the rankings produced by these metrics deviate significantly from human rankings.
In contrast, we introduce VLM-as-a-Judge with reference captions, which demonstrates robust discernment for detailed captions.
It achieves the highest alignment with human judgments at both the caption and model levels.

In light of these findings, we release \caparenaA, an automated benchmark for detailed captioning.
It comprises 600 samples and innovatively adopts the pairwise battle paradigm to improve evaluation reliability.
VLM-as-a-Judge is employed to estimate human preferences by comparing captions against three baseline models.
With 94.3\% correlation to human rankings and just \$4 per test, \caparenaA{} offers a fast and robust pipeline for evaluating detailed captioning.

\begin{table*}[ht]
\centering
\small
\resizebox{1.0\linewidth}{!}{%
\begin{tabular}{p{0.15\textwidth}p{0.7\textwidth}p{0.15\textwidth}}
\toprule
\textbf{Image} & \textbf{Pairwise Caption Battle} & \textbf{Preference} \\
\midrule
\begin{minipage}{.15\textwidth}
\includegraphics[height=25mm]{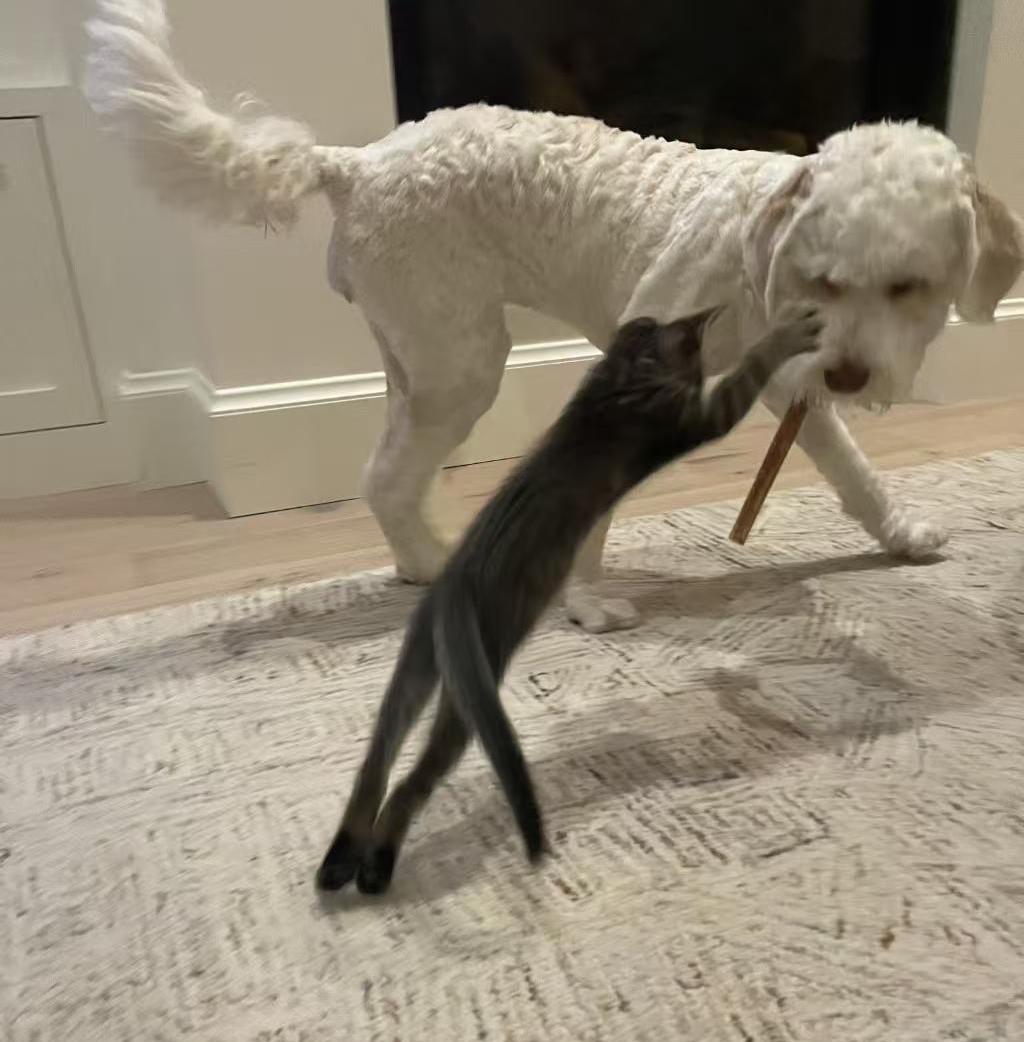}
\end{minipage}
& 
\begin{minipage}{.7\textwidth}
\textbf{Caption1 (Qwen2-VL-72B):} ... The dog is holding a stick in its mouth, and \redmark{the kitten is standing on its hind legs}, reaching up to grab the stick. The kitten's front paws are extended towards the stick, and \redmark{its body is slightly arched} as it tries to take the stick from the dog. ... 
\\
\textbf{Caption2 (Human):} ... A gray and black kitten is \greenmark{leaping into the air at the dog's face}. It is \greenmark{facing the back at a right angle}. Its right leg is extended out with its paw in front of the dog's face. \greenmark{Its tail is down to the right in the air}. ... 
\end{minipage}
& 
\begin{minipage}{.15\textwidth}
    Caption2 is better (more accurate and vivid description of the cat's posture).
\end{minipage}
\\
\midrule
\begin{minipage}{.15\textwidth}
\includegraphics[height=27mm]{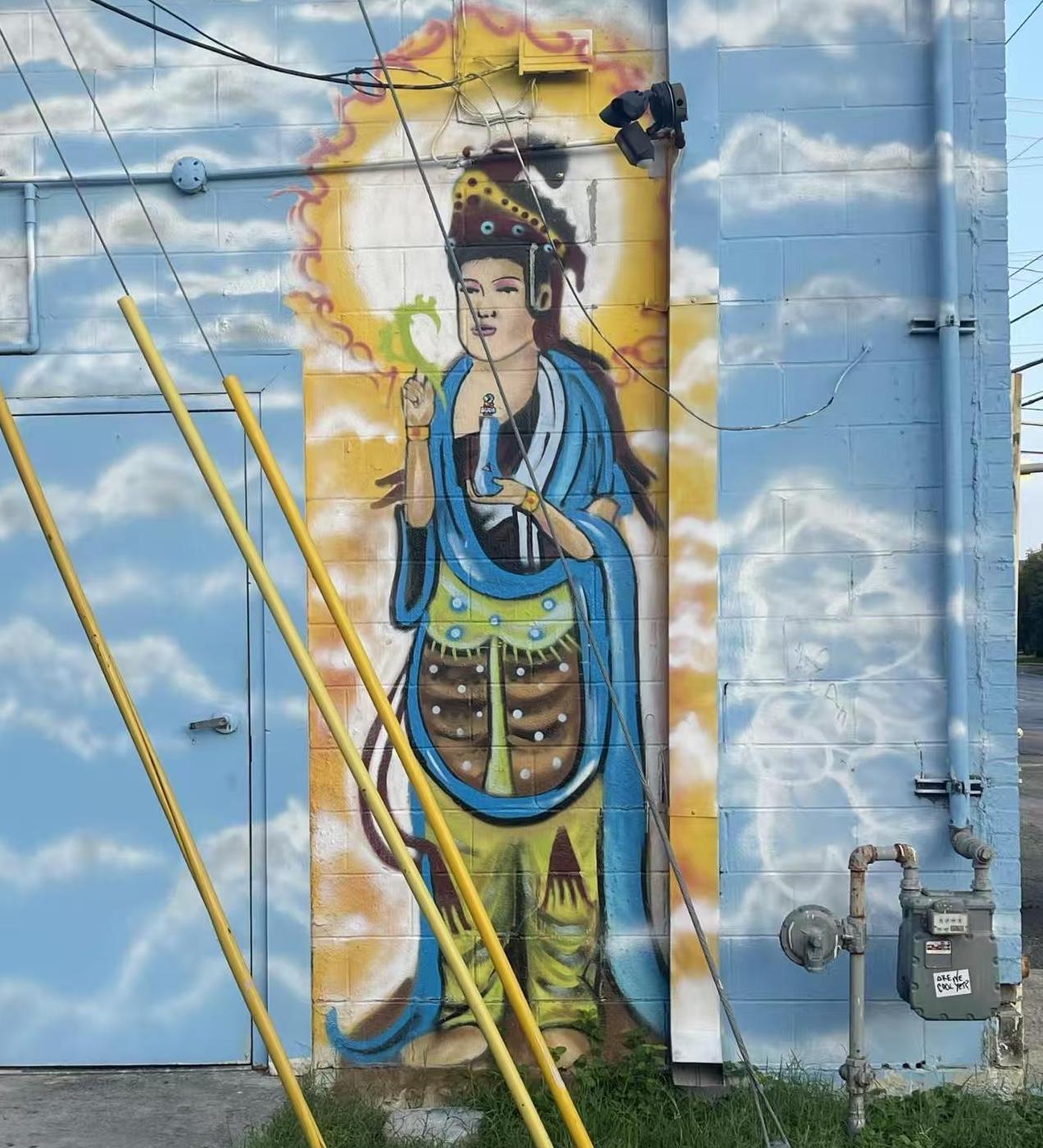}
\end{minipage}
& 
\begin{minipage}{.7\textwidth}
\textbf{Caption1 (GPT-4o):} ... The mural features a figure that appears to be inspired by \greenmark{traditional Asian art, possibly a deity or spiritual figure}. The figure is depicted with a serene expression ... Surrounding the figure is a halo-like glow in warm tones of yellow and orange, \greenmark{adding a sense of divinity or spirituality}.
\\
\textbf{Caption2 (LLama-3.2-90B):} The image depicts \redmark{a vibrant mural of a woman} on the side of a building, surrounded by various objects and features. **Mural:** The mural is painted in bright colors, featuring \redmark{a woman} with long dark hair wearing a blue robe ...
\end{minipage}
& 
\begin{minipage}{.15\textwidth}
    Caption1 is better (\textit{a spiritual figure} is much more informative than \textit{a woman}).
\end{minipage}
\\
\bottomrule
\end{tabular}
}
\caption{Examples of pairwise battles in \caparena. Red and green indicate less accurate and more preferable expressions, respectively. The evaluation guidelines are detailed in \Cref{sec:protocol}, and more examples are in \Cref{tab:full-protocol-example}.}
\label{tab:evaluation-protocol}
\end{table*}

\section{Related Work}

\subsection{Vision-Language Models}
In recent years, Vision-Language Models (VLMs) have experienced rapid advances, achieving state-of-the-art performance across various multimodal tasks. 
Most VLMs integrate visual encoders ~\citep{radford2021learning, zhai2023sigmoid} with large language models ~\citep{bai2023qwen,touvron2023llama}, allowing the latter to possess visual perception capabilities ~\citep[][\emph{inter alia}]{liu2024visual, ye2023mplug, bai2023qwen, chen2024far}. 
To achieve this, VLMs are typically trained in two stages: pre-training on large-scale caption data to align visual and textual information, followed by supervised fine-tuning (SFT) with instruction-following capacity.
Captioning plays a crucial role in building VLMs \cite{chen2024sharegpt4v}, as accurately describing image content forms the foundation for complex recognition and reasoning tasks \cite{cheng2024seeclick}.
Current VLM evaluations mainly focus on visual question answering related to knowledge \cite{yue2024mmmu} and reasoning \cite{wu2024atlas, sun2024genesis}, while captioning is often overlooked due to evaluation difficulties.
Our work is dedicated to benchmarking and analyzing the captioning ability of VLMs.

\subsection{Image Captioning and Evaluation Metrics}
Image captioning significantly progressed over the past decade~\cite{vinyals2015show, xu2015show, anderson2018bottom, cheng2022ads, wang2022ofa, ma2023food, cheng2023beyond, pi2024image}.
These methods use human-annotated datasets like MSCOCO \cite{lin2014microsoft} and Nocaps \cite{agrawal2019nocaps}, evaluating generated captions by comparing them to reference sentences using rule-based metrics such as BLEU \cite{papineni2002bleu} and CIDEr \cite{vedantam2015cider}.
Recent research shows that CLIP-based metrics \cite{hessel2021clipscore} exhibit higher human consistency for short captions.
However, current VLMs generate significantly longer, detailed captions, where traditional metrics are not effective, posing challenges for evaluation.

Several recent works have focused on detailed captioning. \citet{dong2024benchmarking} introduced a new metric CAPTURE and improved VLM captioning performance.
\citet{lu2024benchmarking} proposed a sophisticated scene graph-based metric.
Differently, we first conducted a large-scale human-centered empirical study
to systematically evaluate advanced VLMs.
We then concentrated on the consistency between metrics and human preferences. An in-depth analysis using human annotated data (50 times larger than previous works) reveals the robustness and systematic biases of different metrics.

\section{\caparena: Benchmarking VLMs in Detailed Image Captioning}

While current VLMs excel in tasks like visual perception, question answering, and reasoning, their ability to generate long, detailed image descriptions remains unclear.
In this section, we address this gap by introducing the first large-scale human-centered empirical study to benchmark VLMs in the context of detailed image captioning.

Next, we first present our evaluation protocol tailored for detailed captioning task (\Cref{sec:protocol}).
Then, we describe the implementation of our annotation platform \caparena{} (\Cref{sec:infra}).
Finally, we highlight our key findings on the performance of existing advanced VLMs (\Cref{sec:performance}).

\subsection{Evaluation Protocol}
\label{sec:protocol}

We originally tried a scoring system \cite{hodosh2013framing}, where annotators were asked to assign a score from 1 to 5 to a single description.
However, the task proved to be inherently complex and subjective.
Since most generated captions cover the main content of the image, annotators found it difficult to assign precise grades, leading to low inter-annotator consistency.
Inspired by open-domain evaluations of LLMs, we shifted to a pairwise comparison methodology \cite{chiang2024chatbot} to assess detailed captions, which was further validated in our preliminary study.

We believe a reliable evaluation protocol is crucial for accurate assessments. 
Inspired by \citet{kasai2021transparent}, expert annotators drafted the initial guidelines, which were refined through in-house meetings with all annotators to ensure consistency. 
Finally, we established the following transparent evaluation protocol for detailed captioning.

\noindent
\textbf{Guidlines.}
Our guidelines primarily evaluate the quality of descriptions in terms of precision and informativeness.
The full annotator guidelines can be found in \Cref{app:guideline}.

\noindent
\textbf{Precision}: 
Precision measures how precise the content in the description is, i.e., whether the description aligns with the details in the image.
For example, in the first case of \Cref{tab:evaluation-protocol}, Qwen2-VL provides an inaccurate description of the cat's posture, while the human description captures the crucial action of the cat pouncing toward the dog.
Precision includes various aspects, such as objects, attributes, relationships, and positions.

\noindent
\textbf{Informativeness}:
Informativeness assesses how much of the key information in the image is comprehensively covered by the description, including the salient objects and important details.
For example, in the second case of \Cref{tab:evaluation-protocol}, Llama-3.2's description of \textit{a woman} is precise; however, it is clearly less informative compared to GPT-4o's description of \textit{a spiritual figure}.

\noindent
\textbf{Hallucination}:
Hallucinations are considered an intractable flaw in VLMs \cite{li2023evaluating}, where models generate objects that do not exist in the image. 
We instruct annotators to impose strict penalties for hallucinations, as they significantly harm the caption quality in real-world applications.

Additionally, previous studies show that human annotators are influenced by output length and response style \cite{chiang2024chatbot}.
To mitigate this, we asked annotators to focus solely on the quality of the descriptions, minimizing distractions from such aspects.
Furthermore, for pairs of similar quality, we considered the longer description less favorable if it was noticeably too long.

\subsection{\caparena: Pairwise Battle Platform}
\label{sec:infra}
We developed \caparena, an annotation platform aimed at benchmarking VLMs' performance in detailed captioning through anonymous pairwise battles.
The platform covers a diverse range of image scenarios and evaluates a set of established VLMs through human annotator votes, providing reliable rankings between models.

\noindent
\textbf{Data Source.}
The test images are sourced from the recently proposed DOCCI dataset \cite{onoe2024docci}, which includes high-resolution images of various real-life scenes.
We provide detailed information about this dataset in \Cref{app:docci}.
Notably, each image in this dataset is paired by carefully crafted long, human-annotated descriptions, which are used as the human baseline on our platform.

\begin{figure}[t]
  \centering
  \includegraphics[width=0.95\columnwidth]{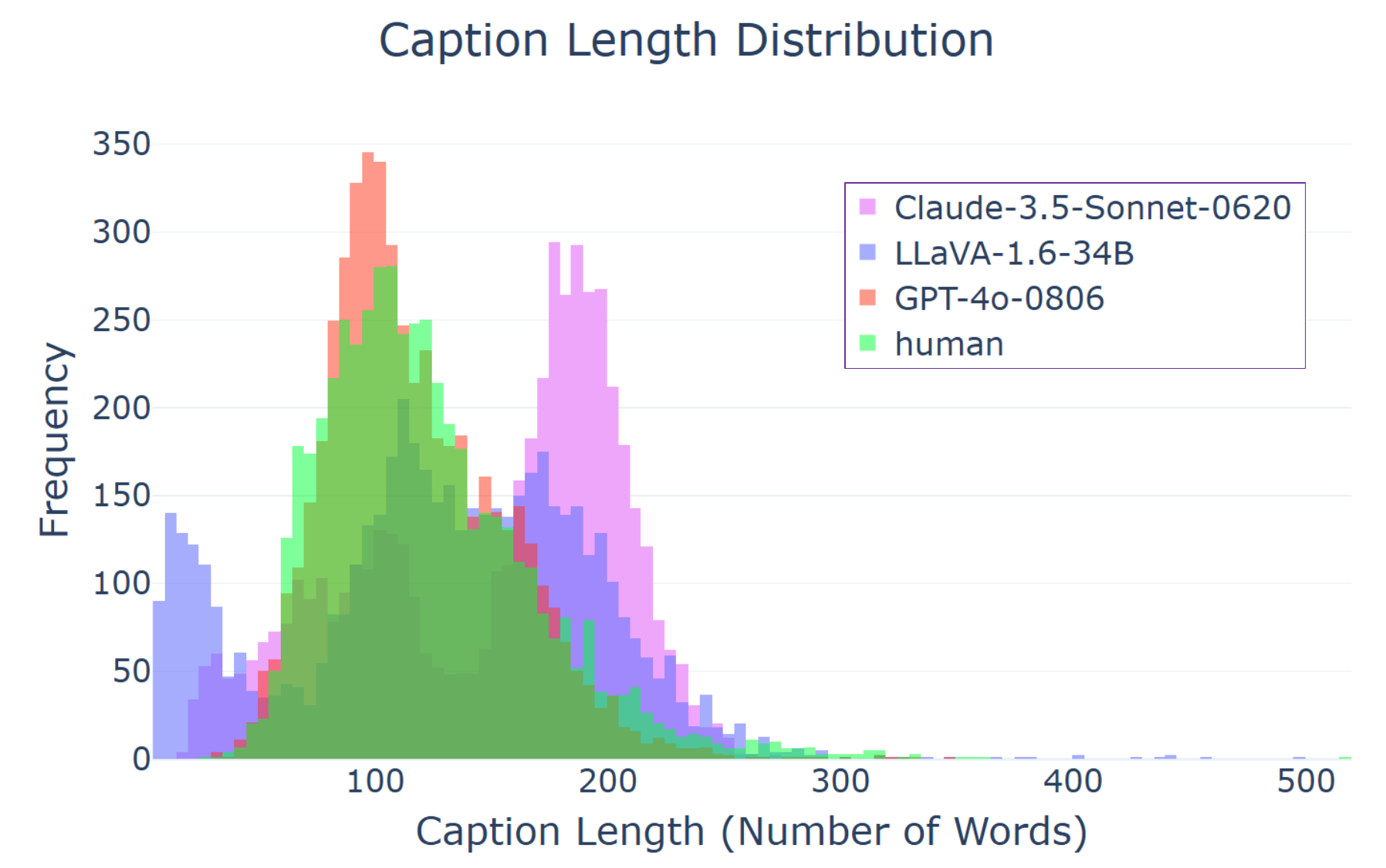} 
  \caption{Caption length distribution of different VLMs.}
  \label{fig:length}
\end{figure}

For the tested models, we selected a diverse set of representative VLMs, including both commercial and open-source models, spanning a range of model sizes (the full list is provided in \Cref{app:model}).
To minimize bias introduced by specific prompts, we crafted 10 prompts for detailed image captioning, such as \textit{Describe this image in detail}.
These prompts were manually reviewed to ensure they generated descriptions of similar quality and length (all prompts are provided in \Cref{app:prompt}).
We show the caption length distribution of different VLMs in \Cref{fig:length}, where GPT-4o is most similar to humans.

\noindent
\textbf{\caparena{} Infrastructure.} 
We take $N$ to denote the number of models. Let $n_1, n_2 \in [N]$ be the indices of the models. We define $A_t = (n_1, n_2)$ as the model pair compared at time $t$. The human response is indicated by $H_t \in \{0, 1\}$, where $H_t = 0$ indicates human preference for model $n_1$, and $H_t = 1$ for model $n_2$.

To focus on comparing models with similar performance levels and thus accelerate the convergence of rankings, we adopt the probability update strategy from Chatbot Arena \cite{chiang2024chatbot}.
The sampling probability for each pair is proportional to the reduction in the size of the confidence interval:
\[
P_t(a) \propto \sqrt{\frac{\hat{\Sigma}_{t, a, a}}{|\{ t : A_t = a \}|}} - \sqrt{\frac{\hat{\Sigma}_{t, a, a}}{|\{ t : A_t = a \}|+ 1} },
\]
where $P_t(a)$ is the probability of sampling pair $a$ at time $t$, and $\hat{\Sigma}_t$ denotes the covariance matrix estimated from the $t$ samples.

We then apply the Bradley-Terry (BT) model \cite{BT} with the logistic form to calculate the scores of the models:
\[
\hat{s} = \arg\min_{\xi} \sum_{t=1}^{T} \frac{1}{P(A_t)} \ell \left( H_t, \frac{1}{1 + e^{\xi_{A_{t, 2}} - \xi_{A_{t, 1}}}} \right),
\]
where $\xi$ is the vector of BT coefficients, which is an $N$-dimensional vector, and $\ell$ is the binary cross-entropy loss, defined as $\ell(h, p) = -(h \log(p) + (1 - h) \log(1 - p))$. 
The BT coefficients $\hat{s}$ are used to create the ordered ranking of models. We bootstrap the BT rating estimate 1000 times to construct a confidence interval for each rating. 

\begin{figure*}[t]
  \centering
  \includegraphics[width=1.0\textwidth]{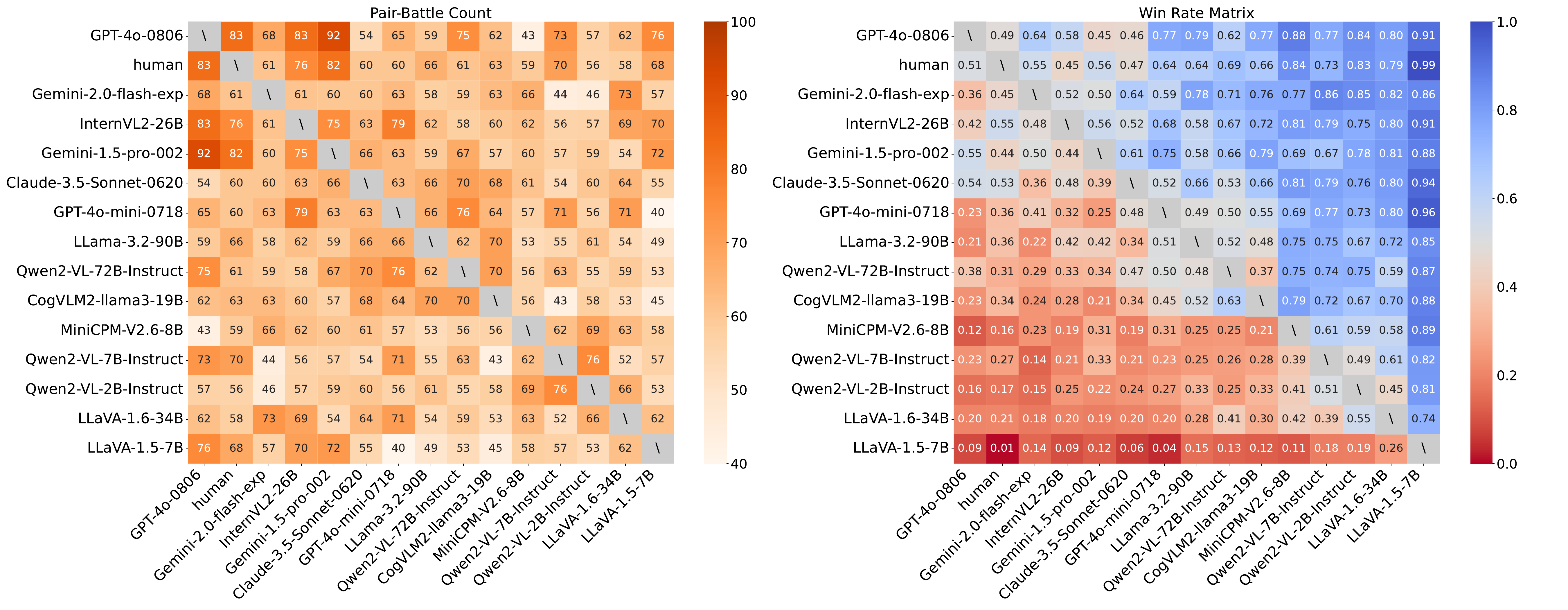} 
  \caption{Battle counts and win rate matrix between models in \caparena.
  }
  \label{fig:bcnwr}
\end{figure*}

\begin{figure*}[t]
        \centering
        \includegraphics[width=0.9\textwidth]{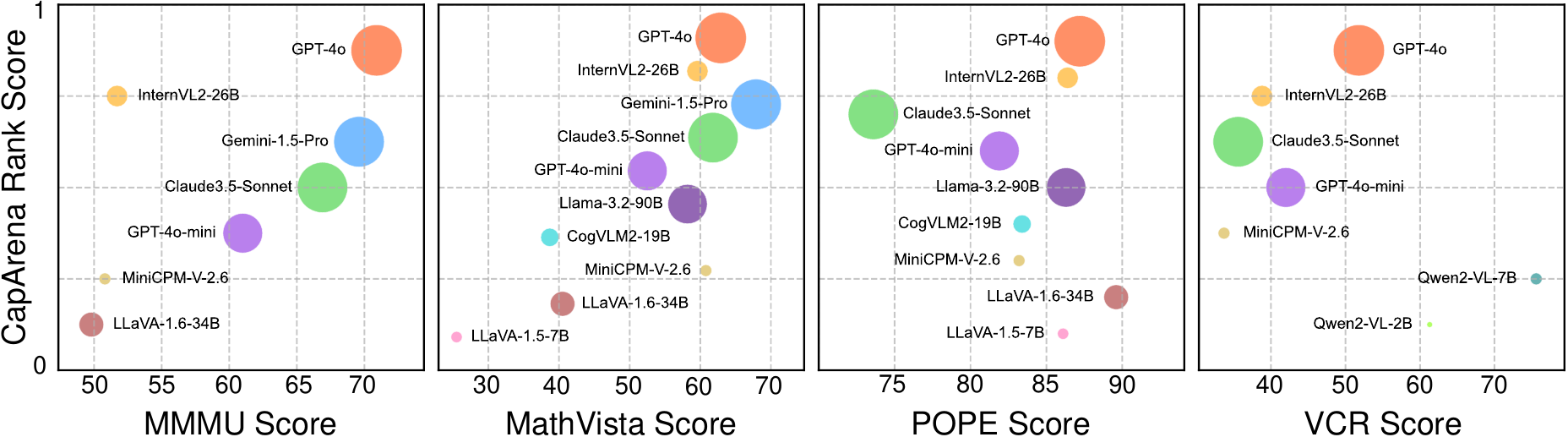}
    \caption{Correlation between vision-language benchmark scores and CapArena ranking. Models that perform well on general benchmarks do not necessarily excel in captioning. The size of each point represents the model size.}
    \label{fig:cor_bench}
\end{figure*}

\noindent
\textbf{Annotator Training and Quality Control} 
We made considerable efforts to enhance the reliability of the annotations.
We conducted in-house annotations, with all annotators being graduate students specializing in natural language processing, with no visual impairments, and familiar with the image captioning task.\footnote{The first four authors of participated in the annotation.} 
We first conducted a preliminary annotation on 100 captions and developed the initial guidelines. In-house workshops followed to refine the process and ensure all annotators understood the task.
Finally, we adopted the annotation methodology described in Section \ref{sec:protocol}.

During the annotation process, we randomly selected 400 samples to be annotated by different annotators for inter-annotator validation. 
The results showed an agreement of 0.782, demonstrating the reliability of our manual labeling.
Our empirical study found that the remaining 20\% of disagreement stemmed not from annotator errors, but from the inherent subjectivity in interpreting which aspects of the image to emphasize.
We also implemented measures to continuously monitor annotation quality during the labeling process (see \Cref{app:qualitycontrol}).
Data collection began in October 2024, and by February 2025, a total of 6,522 annotation instances had been collected.
The average time per annotation was 142 seconds.

\subsection{VLMs Performance Analysis}
\label{sec:performance}

The final ranking is shown in \Cref{fig:ranking}.
\Cref{fig:bcnwr} presents the battle count and win rate matrix.

\paragraph{Top models achieve human-level performance.}
Surprisingly, our evaluation reveals for the first time that leading models like GPT-4o have reached or even surpassed human-level performance in detailed captioning tasks.
This milestone demonstrates that machine-generated descriptions can now rival high-quality human descriptions.
We conducted an empirical study in \Cref{app:GPTvsHuman}, and the results show that even human expert annotators occasionally overlook image details, where GPT-4o's descriptions are more comprehensive.

\paragraph{\caparena{} uncovers disparities in fine-grained visual perception across models.}
As shown in \Cref{fig:ranking}, most open-source models still lag significantly behind commercial models, even with large model sizes (e.g., Llama-3.2-90B).
Despite competitive performance on general multimodal benchmarks, our results reveal the limitations of open-source alternatives in detailed captioning, where they lack stability in accurately perceiving image details.
An exception is InternVL2-26B, a mid-sized model that demonstrates exceptional performance.
We attribute this to its large-scale vision encoder, InternViT-6B, enabling strong visual understanding.
These results highlight the growth potential of open-source models.

We present the correlation between model performance on vision-language benchmarks and CapArena ranking in \Cref{fig:cor_bench}. 
Strong performance on general benchmarks does not always translate to superior captioning ability.
For example, despite its relatively low MMMU \cite{yue2024mmmu} score, InternVL2-26B excels in captioning.
Moreover, the hallucination benchmark POPE \cite{li2023evaluating} lacks discriminative capacity for detailed captioning, underscoring the need for further exploration in assessing image comprehension.

\paragraph{Failure Cases.}
Besides weaker models missing image details or making errors, we identified several common mistakes:
(1) failure to provide accurate descriptions for unusual scenes, 
(2) frequent neglect of subtle yet important details,
(3) challenges in scenes requiring knowledge association,
and (4) an inability of VLMs to accurately identify the time on clocks.
Examples of these cases are provided in \Cref{fig:failure-case}.
These shortcomings highlight the need for further research in detailed captioning to advance real-world applications.

\section{Analysis of Captioning Metrics}

Thanks to its meticulously designed annotation process, \caparena{} can be regarded as a reliable evaluation system for detailed image captioning that reflects the performance of different models.
However, it relies on costly human preference annotations, making annotating every model of interest impractical.
Therefore, automatic metrics are crucial for evaluating and iterating on the development of detailed captioning capabilities in VLMs.

This section evaluates a range of traditional and recently proposed captioning metrics, and the ability of VLM-as-a-Judge to assess the quality of detailed captions using the \caparena{} data.
The over 6000 high-quality human annotations serve as the golden standard, providing the basis to analyze how these metrics correlate with human preferences and to identify their potential deficiencies.

\subsection{Captioning Metrics}

\noindent
\textbf{Traditional Metrics.}
We first consider rule-based metrics that are commonly used for MSCOCO evaluation:
\textbf{BLEU} \cite{papineni2002bleu} measures n-gram overlap between the generated and reference captions.
\textbf{METEOR} \cite{banerjee2005meteor} incorporates synonym matching and stemming to improve recall.
\textbf{SPICE} \cite{anderson2016spice} measures semantic consistency using scene graph.
\textbf{CIDEr} \cite{vedantam2015cider} evaluates consensus across multiple reference captions.

Next, we include a range of CLIP-based metrics, which claim higher alignment with human judgments \cite{hodosh2013framing}.
\textbf{CLIPScore} \cite{hessel2021clipscore} is a reference-free metric that measures the similarity using CLIP features, and \textbf{LongCLIPScore} \cite{zhang2024long} is a variant adapted for longer text.
\textbf{Polos} \cite{wada2024polos} enhances CLIPScore through supervised learning on human feedback.
\textbf{FLEUR} \cite{lee2024fleur} leverages a large multimodal model and achieves state-of-the-art human alignment on short captions.

\noindent
\textbf{Metrics Designed for Detailed Captions.}
To evaluate long, detailed descriptions, a few specialized metrics have been proposed.
\textbf{CAPTURE} \cite{dong2024benchmarking} extracts visual objects, attributes, and relationships, performing multi-stage matching between generated and reference captions.
\textbf{VDCScore} \cite{chai2024auroracap} is a video captioning metric that decomposes the reference caption into question-answer pairs and utilizes an LLM to check their correspondence with the predicted caption.

\noindent
\textbf{VLM-as-a-Judge.}
Leveraging powerful LLMs to simulate human preferences has proven effective in open-ended scenarios \cite{zheng2023judging}.
While recent studies have explored VLM-as-a-Judge in general multimodal tasks \cite{chen2024mllm, li2024vlrewardbench}, we are the first to apply it to detailed caption comparison—a task that demands fine-grained discernment of vision-language semantic alignment.
Similar to human judgment, given an image and two descriptions, VLMs determine which one is better.
We employ several well-established VLMs (e.g., GPT-4o, Qwen2.5-VL \cite{bai2025qwen2}, LLaVA-OneVision \cite{li2024llava}, LLaVA-Critic \cite{xiong2024llava}) as evaluator.
We also introduce a reference-enhanced variant that incorporates human descriptions to assist the model's judgment.

\begin{table*}[ht]
\centering
\footnotesize
\renewcommand\arraystretch{1.1}
\resizebox{0.95\linewidth}{!}{%
\begin{tabular}{lcccccccc}
\toprule
\multicolumn{1}{l|}{}  & \multicolumn{1}{c|}{\textbf{Need}} & \multicolumn{5}{c|}{\textbf{Caption-level Agreement (Including Tie)}} & \multicolumn{2}{c}{\textbf{Model-level Agreement}} \\
\multicolumn{1}{l|}{\multirow{-2}{*}{\textbf{Metrics}}} & \multicolumn{1}{c|}{\textbf{Ref?}} & \textbf{Overall} & \textbf{Level1} & \textbf{Level2} & \textbf{Level3} & \multicolumn{1}{c|}{\textbf{Level4}} & \multicolumn{1}{c}{\textbf{Spearman}} & \multicolumn{1}{c}{\textbf{Kendall $\tau$}} \\
\midrule
\multicolumn{1}{l|}{Inter-Annotator}  & \multicolumn{1}{c|}{-}   & 0.683 & 0.810 & 0.650 & 0.650 & \multicolumn{1}{c|}{0.620} & -    & -     \\
\multicolumn{1}{l|}{Output Length}     & \multicolumn{1}{c|}{No}  & 0.585 & 0.672 & 0.593 & 0.552 & \multicolumn{1}{c|}{0.521} & 0.710 & 0.582 \\
\midrule
\multicolumn{9}{c}{\cellcolor{gray!25}\textbf{Traditional Image Captioning Metrics}} \\
\midrule
\multicolumn{1}{l|}{BLEU-4}           & \multicolumn{1}{c|}{Yes} & 0.474 & 0.477 & 0.480 & 0.475 & \multicolumn{1}{c|}{0.467} & 0.424    & 0.319   \\
\multicolumn{1}{l|}{SPICE}            & \multicolumn{1}{c|}{Yes} & 0.417 & 0.441 & 0.422 & 0.415 & \multicolumn{1}{c|}{0.387} & 0.275      & 0. 231     \\
\multicolumn{1}{l|}{CIDER}            & \multicolumn{1}{c|}{Yes} & 0.384 & 0.378 & 0.383 & 0.389 & \multicolumn{1}{c|}{0.387} &    -0.279   & -0.209      \\
\multicolumn{1}{l|}{METEOR}           & \multicolumn{1}{c|}{Yes} & 0.576 & 0.657 & 0.582 & 0.536 & \multicolumn{1}{c|}{0.530} & 0.785 & 0.582 \\
\multicolumn{1}{l|}{Polos}            & \multicolumn{1}{c|}{Yes} & 0.479 & 0.526 & 0.467 & 0.462 & \multicolumn{1}{c|}{0.462} &    0.420   & 0.363     \\
\multicolumn{1}{l|}{CLIPScore}         & \multicolumn{1}{c|}{No}  & 0.325 & 0.266 & 0.308 & 0.362 & \multicolumn{1}{c|}{0.355} &   -0.574    & -0.451     \\
\multicolumn{1}{l|}{LongCLIPScore}     & \multicolumn{1}{c|}{No}  & 0.400 & 0.422 & 0.404 & 0.384 & \multicolumn{1}{c|}{0.395} &    -0.226   & -0.121     \\
\multicolumn{1}{l|}{FLEUR}             & \multicolumn{1}{c|}{No}  & 0.458 & 0.513 & 0.462 & 0.444 & \multicolumn{1}{c|}{0.414} &   0.393  & 0.297   \\
\midrule
\multicolumn{9}{c}{\cellcolor{gray!25}\textbf{Metrics Designed for Detailed Image Captioning}} \\
\midrule
\multicolumn{1}{l|}{CAPTURE}          & \multicolumn{1}{c|}{Yes}  & 0.525 & 0.601 & 0.512 & 0.504 & \multicolumn{1}{c|}{0.479} &   0.613  & 0.538   \\
\multicolumn{1}{l|}{VDC-Score}        & \multicolumn{1}{c|}{Yes}  & 0.557 & 0.687 & 0.579 & 0.496 & \multicolumn{1}{c|}{0.460} &      0.890 &   0.736   \\
\midrule
\multicolumn{9}{c}{\cellcolor{gray!25}\textbf{VLM-as-a-Judge}} \\
\midrule
\multicolumn{1}{l|}{LLaVA-OneVision}       & \multicolumn{1}{c|}{No}    & 0.606      & 0.709      & 0.642      & 0.541      &  \multicolumn{1}{c|}{0.537}     & 0.921      & 0.780     \\
\multicolumn{1}{l|}{LLaVA-Critic}       & \multicolumn{1}{c|}{No}    & 0.609      & 0.735      & 0.631      & 0.544      &  \multicolumn{1}{c|}{0.530}     & 0.903      & 0.736     \\
\multicolumn{1}{l|}{Qwen2.5-VL}       & \multicolumn{1}{c|}{No}    & 0.625      & 0.739      & 0.647      & 0.566      &  \multicolumn{1}{c|}{0.552}     & 0.908      & 0.736     \\
\multicolumn{1}{l|}{GPT-4o}            & \multicolumn{1}{c|}{No}  & \textbf{0.628} & \textbf{0.740} & 0.647 & \textbf{0.572} & \multicolumn{1}{c|}{0.557} & 0.930      & 0.802      \\
\multicolumn{1}{l|}{GPT-4o (with ref)} & \multicolumn{1}{c|}{Yes} & 0.627 & 0.733 & \textbf{0.663} & 0.559 & \multicolumn{1}{c|}{\textbf{0.560}} & \textbf{0.943} & \textbf{0.846} \\
\bottomrule
\end{tabular}
}
\caption{Evaluation of various metrics on detailed captioning tasks. \textit{Need Ref} indicates whether human-written reference captions are required. Higher scores indicate better alignment with human preferences. The best results in each column are highlighted in \textbf{bold}. GPT-4o as the evaluator achieves the best performance.}
\label{tab:captioning_metrics}
\end{table*}

\subsection{Experiment Settings}

To evaluate these metrics' ability to assess detailed caption quality, we compare metric-based judgments with human annotations on caption battle pairs in \caparena{} and analyze their correlation with human preferences. 
For scoring-based metrics like CIDEr and CLIPScore, scores are used to determine the winner in pairwise battles.
For VLM-as-a-Judge, we use a prompt similar to human annotator guidelines (See \Cref{app:templete}).
Since some pairs are of similar quality, we allow ties to align with human annotations.
Scoring-based metrics use a threshold to determine draws, ensuring a similar occurrence rate as in human annotations. 

We analyze the agreement between metrics and human preferences from two perspectives:

\noindent
\textbf{Caption-level Agreement.}
Caption-level agreement measures the consistency between metrics and human judgments for the same pairwise caption battle. It is calculated as the proportion of pairs in \caparena{} where the metric's decision matches the human judgment (A wins / B wins / tie).
To assess metric performance across caption pairs of varying distinction, we categorize all samples into four levels based on the ranking gap between models.
Level 1 consists of the most easily distinguishable model pairs (e.g., GPT-4o vs. LLaVA-1.5), while Level 4 includes those with the most similar performance (e.g., GPT-4o vs. Gemini-1.5).

We also report human annotators' internal consistency across the four levels for reference, manually verifying 100 samples per level.

\noindent
\textbf{Model-level Agreement.}
Caption-level agreement measures how often a metric matches human judgments on individual battles.
However, a metric may achieve high caption-level agreement while exhibiting bias toward certain models, leading to inaccurate performance estimation.
To address this, we introduce model-level agreement, which measures the consistency between rankings derived from human judgments and those derived from metrics. 
Specifically, we replace human annotations with metrics for all pairwise battles in \caparena{} and compute rankings using the same ELO mechanism.
We compute this agreement using Spearman’s rank coefficient and Kendall’s $\tau$.

\begin{figure*}[ht]
\centering
\includegraphics[width=0.9\textwidth]{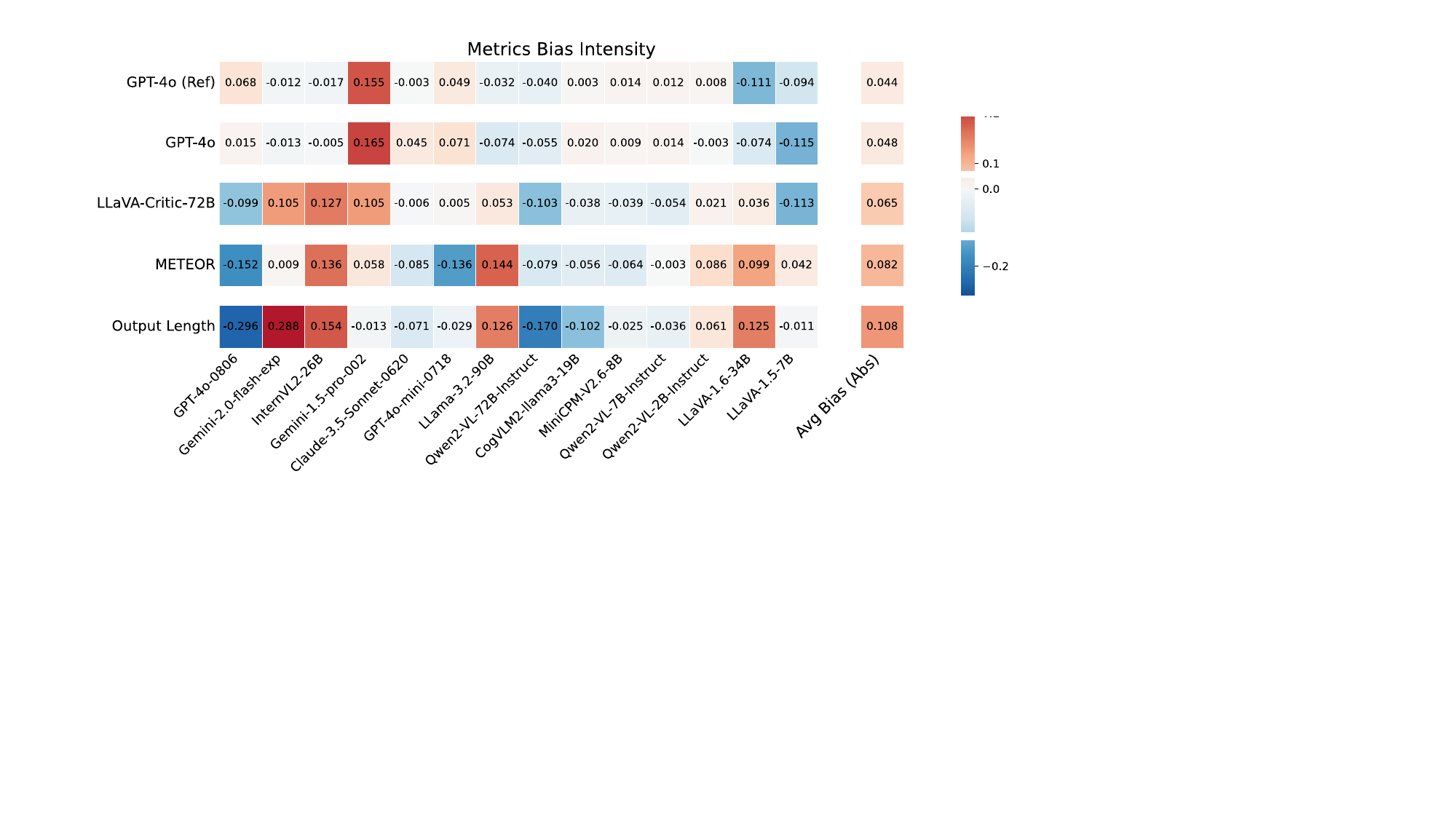}
\caption{Metrics exhibit systematic biases—overestimating (\textcolor{darkred}{positive values}) or underestimating (\textcolor{darkblue}{negative values}) certain models. Color saturation represents bias magnitude. Different metrics favor different models. GPT-4o exhibits lower biases (lighter overall colors), contributing to higher model-level agreement observed in \Cref{tab:captioning_metrics}.}
\label{fig:bias}
\vspace{-1.0em}
\end{figure*}

\subsection{Results Analysis}

\paragraph{Most traditional metrics fail in the detailed captioning task.}
Traditional metrics with high human agreement on short captions \cite{hodosh2013framing} deviate from human preferences when applied to detailed captioning task.
As shown in \Cref{tab:captioning_metrics}, they exhibit low agreement both at the caption-level and model-level.
Rule-based metrics, such as CIDEr, struggle due to the flexible nature of detailed captions, which complicates n-gram matching.
For CLIP-based metrics like CLIPScore, our results reveal that current vision-language representation models fail to align fine-grained details in long descriptions to the image content.

\paragraph{VLM-as-a-Judge demonstrates stronger discernment for detailed captions.}
As shown in \Cref{tab:evaluation-protocol}, accurate judgment for closely matched battles relies on a deep understanding of image content and a discriminative comparison between descriptions.
Powerful VLMs with reasoning and fine-grained perception exhibit this capacity. 
No bells and whistles, GPT-4o as an evaluator, shows the highest consistency with human preferences, outperforming recently proposed metrics.
Reference descriptions help the evaluator clarify uncertain image details, further improving model-level agreement.

\paragraph{Systematic biases—overestimating or underestimating certain models are a critical concern.}
As shown in \Cref{tab:captioning_metrics}, METEOR and Output Length (which simply favors longer captions) exhibit decent caption-level agreement, but their model-level agreement is notably lower.
To investigate this issue, we assessed the metric's intrinsic bias towards specific models, by calculating the model's average win rate across all battles and comparing it with the golden win rate in \Cref{fig:bcnwr}. 
A positive (negative) win rate difference indicates that the metric overestimates (underestimates) the model.
As displayed in \Cref{fig:bias}, all metrics exhibit systematic biases. 
The degree of bias varies across metrics; Output Length shows a particularly strong bias, while GPT-4o-as-a-Judge has a lower bias than METEOR (average 4.4\% vs. 8.2\%).
This suggests that the disagreement between GPT-4o-as-a-Judge and humans is more likely due to random preferences per independent sample, rather than harmful bias towards specific models, leading to more accurate model ranking estimates.

\paragraph{Evaluating hard-to-distinguish caption pairs remains a challenge.}
Despite the reasonable performance of VLM-as-a-Judge, it still falls short of Inter-Annotator agreement, particularly on Level 3/4 samples which represent the most challenging model battles (e.g., 0.650/0.620 vs. 0.572/0.560).
This suggests that the current VLM still has limitations in perceiving fine-grained image details and distinguishing subtle differences between captions, which in turn leaves room for further improvement in detailed caption evaluation.

\section{\caparenaA: An Automated Benchmark for Detailed Captioning}
\caparena{} relies on substantial human labor to estimate model performance, which is time-consuming and expensive. 
Therefore, developing an automated benchmark for detailed captioning is a desideratum to enable rapid evaluation and accelerate model development.
In this section, based on the findings above, we introduce \caparenaA, a cheap and fast framework for detailed captioning evaluation. 
\caparenaA{} includes 600 evaluation images and assesses model performance through pairwise battles with baseline models.

\paragraph{Curation of Test Samples.}
We selected images from the DOCCI test split \cite{onoe2024docci} as candidates.
The images, photographed by diverse annotators in everyday contexts, encompass a wide range of scenes and are of high resolution.
Given that these images were newly captured and released after the timestamp of most existing VLMs’ training, we believe there is minimal risk of data contamination.
Next, to sample a diverse test set, we adopted the image feature clustering from DOCCI, uniformly sampling 600 samples from 149 clusters. 
Additionally, we applied CLIP feature-based filtering to remove overly similar samples, ensuring the quality of the final selection.

\paragraph{Evaluation Protocol.}
We employ a pairwise battle paradigm and use VLM-as-a-Judge for comparison \cite{li2024crowdsourced, chou2024visionarena}.
For each of the 600 test samples, we compare captions from the test model and a baseline model to determine the better one. 
To reduce potential noise and bias from a single baseline \cite{lin2024wildbench}, we use three baseline models with different performance levels: GPT-4o, CogVLM-19B, and MiniCPM-8B.
We use GPT-4o as the judge due to its high agreement with human preferences and provide human reference captions as additional support.

To compute the final score for the test model, we assign +1 for a win, -1 for a loss, and 0 for a draw in each pairwise comparison.
The model's score in \caparenaA{} is the total sum of its scores across the 600 test samples. 
We provide the current leaderboard of \caparenaA{} in \Cref{tab:caparena-auto}.

\begin{table}[ht]
\centering
\resizebox{1.0\linewidth}{!}{%
\begin{tabular}{lcc}
\toprule
                    & \textbf{Spearman} & \textbf{Kendall $\tau$} \\
\midrule
DOCCI (with BLEU-4)  & 0.341             & 0.275                     \\
DOCCI (with METEOR) & 0.859             & 0.648                \\
CAPTURE             & 0.763             & 0.604                \\
\caparenaA          & \textbf{0.943}             & \textbf{0.824}                \\
\bottomrule
\end{tabular}
}
\caption{Correlation between the automated benchmark and \caparena’s golden ranking. \caparenaA{} exhibits the highest alignment with human preferences.}
\label{tab:caparena_auto}
\vspace{-0.5em}
\end{table}

To validate the effectiveness of \caparenaA, we compare it with several recent detailed captioning benchmarks. 
\Cref{tab:caparena_auto} presents the correlation coefficients between the model rankings from these benchmarks and the golden ranking provided by \caparena.
\caparenaA{} outperforms existing benchmarks by a large margin, better aligning with human preferences.
Additionally, its streamlined design allows each evaluation to cost only \$4, making it an effective and affordable evaluation benchmark for detailed captioning.

\section{Conclusion}
In this paper, we explore the task of detailed image captioning in the LLM era.
We conducted the first large-scale human-centered empirical study to benchmark advanced VLMs performance.
The results demonstrate that, for the first time, leading models (e.g., GPT-4o) achieve or surpass human-level performance, while open-source models lag behind.
Next, we provided an in-depth analysis of existing captioning metrics.
Our experiments reveal that VLM-as-a-Judge sets a new standard, highlighting that systematic biases in existing metrics are the key factor affecting alignment with human preferences.
Finally, we release \caparenaA, an automated benchmark that closely aligns with human preferences, offering a cost-effective and efficient tool for detailed captioning evaluation.
Our findings lay the foundation for the future development of image captioning.

\section*{Limitations}
One limitation of \caparena{} is the scope of models and the domains covered.
Current pairwise battle evaluation includes 14 representative Vision-Language Models (VLMs), but this selection is constrained by the available annotation resources. As a result, several recent models released after our dataset was compiled have not been considered in the evaluation.
Additionally, the images used mostly focus on everyday life scenarios, which means that other domains, such as artwork or medical images, are not yet represented. Extending the evaluation to these domains could provide a more complete picture of VLM performance.



\newpage

\bibliography{main}

\begin{thebibliography}{51}
\providecommand{\natexlab}[1]{#1}

\bibitem[{Agrawal et~al.(2019)Agrawal, Desai, Wang, Chen, Jain, Johnson, Batra, Parikh, Lee, and Anderson}]{agrawal2019nocaps}
Harsh Agrawal, Karan Desai, Yufei Wang, Xinlei Chen, Rishabh Jain, Mark Johnson, Dhruv Batra, Devi Parikh, Stefan Lee, and Peter Anderson. 2019.
\newblock \href {http://openaccess.thecvf.com/content_ICCV_2019/papers/Agrawal_nocaps_novel_object_captioning_at_scale_ICCV_2019_paper.pdf} {Nocaps: Novel object captioning at scale}.
\newblock In \emph{Proceedings of the IEEE/CVF international conference on computer vision}, pages 8948--8957.

\bibitem[{Anderson et~al.(2016)Anderson, Fernando, Johnson, and Gould}]{anderson2016spice}
Peter Anderson, Basura Fernando, Mark Johnson, and Stephen Gould. 2016.
\newblock \href {https://arxiv.org/pdf/1607.08822} {Spice: Semantic propositional image caption evaluation}.
\newblock In \emph{Computer Vision--ECCV 2016: 14th European Conference, Amsterdam, The Netherlands, October 11-14, 2016, Proceedings, Part V 14}, pages 382--398. Springer.

\bibitem[{Anderson et~al.(2018)Anderson, He, Buehler, Teney, Johnson, Gould, and Zhang}]{anderson2018bottom}
Peter Anderson, Xiaodong He, Chris Buehler, Damien Teney, Mark Johnson, Stephen Gould, and Lei Zhang. 2018.
\newblock \href {https://openaccess.thecvf.com/content_cvpr_2018/papers/Anderson_Bottom-Up_and_Top-Down_CVPR_2018_paper.pdf} {Bottom-up and top-down attention for image captioning and visual question answering}.
\newblock In \emph{Proceedings of the IEEE conference on computer vision and pattern recognition}, pages 6077--6086.

\bibitem[{Bai et~al.(2023)Bai, Bai, Yang, Wang, Tan, Wang, Lin, Zhou, and Zhou}]{bai2023qwen}
Jinze Bai, Shuai Bai, Shusheng Yang, Shijie Wang, Sinan Tan, Peng Wang, Junyang Lin, Chang Zhou, and Jingren Zhou. 2023.
\newblock \href {https://arxiv.org/pdf/2308.12966} {Qwen-vl: A frontier large vision-language model with versatile abilities}.
\newblock \emph{arXiv preprint arXiv:2308.12966}.

\bibitem[{Bai et~al.(2025)Bai, Chen, Liu, Wang, Ge, Song, Dang, Wang, Wang, Tang et~al.}]{bai2025qwen2}
Shuai Bai, Keqin Chen, Xuejing Liu, Jialin Wang, Wenbin Ge, Sibo Song, Kai Dang, Peng Wang, Shijie Wang, Jun Tang, et~al. 2025.
\newblock \href {https://arxiv.org/pdf/2502.13923} {Qwen2. 5-vl technical report}.
\newblock \emph{arXiv preprint arXiv:2502.13923}.

\bibitem[{Banerjee and Lavie(2005)}]{banerjee2005meteor}
Satanjeev Banerjee and Alon Lavie. 2005.
\newblock \href {https://aclanthology.org/W05-0909.pdf} {Meteor: An automatic metric for mt evaluation with improved correlation with human judgments}.
\newblock In \emph{Proceedings of the acl workshop on intrinsic and extrinsic evaluation measures for machine translation and/or summarization}, pages 65--72.

\bibitem[{Bradley and Terry(1952)}]{BT}
Ralph~Allan Bradley and Milton~E Terry. 1952.
\newblock \href {https://www.jstor.org/stable/2334029} {Rank analysis of incomplete block designs: I. the method of paired comparisons}.
\newblock \emph{Biometrika}, 39(3/4):324--345.

\bibitem[{Chai et~al.(2024)Chai, Song, Du, Meng, Madhavan, Bar-Tal, Hwang, Xie, and Manning}]{chai2024auroracap}
Wenhao Chai, Enxin Song, Yilun Du, Chenlin Meng, Vashisht Madhavan, Omer Bar-Tal, Jeng-Neng Hwang, Saining Xie, and Christopher~D Manning. 2024.
\newblock \href {https://arxiv.org/pdf/2410.03051} {Auroracap: Efficient, performant video detailed captioning and a new benchmark}.
\newblock \emph{arXiv preprint arXiv:2410.03051}.

\bibitem[{Chen et~al.(2024{\natexlab{a}})Chen, Chen, Zhang, Liu, Wang, Zhou, Zhang, Wan, Zhou, and Sun}]{chen2024mllm}
Dongping Chen, Ruoxi Chen, Shilin Zhang, Yinuo Liu, Yaochen Wang, Huichi Zhou, Qihui Zhang, Yao Wan, Pan Zhou, and Lichao Sun. 2024{\natexlab{a}}.
\newblock \href {https://arxiv.org/pdf/2402.04788} {Mllm-as-a-judge: Assessing multimodal llm-as-a-judge with vision-language benchmark}.
\newblock \emph{arXiv preprint arXiv:2402.04788}.

\bibitem[{Chen et~al.(2024{\natexlab{b}})Chen, Li, Dong, Zhang, He, Wang, Zhao, and Lin}]{chen2024sharegpt4v}
Lin Chen, Jinsong Li, Xiaoyi Dong, Pan Zhang, Conghui He, Jiaqi Wang, Feng Zhao, and Dahua Lin. 2024{\natexlab{b}}.
\newblock \href {https://arxiv.org/pdf/2311.12793} {Sharegpt4v: Improving large multi-modal models with better captions}.
\newblock In \emph{European Conference on Computer Vision}, pages 370--387. Springer.

\bibitem[{Chen et~al.(2024{\natexlab{c}})Chen, Wang, Tian, Ye, Gao, Cui, Tong, Hu, Luo, Ma et~al.}]{chen2024far}
Zhe Chen, Weiyun Wang, Hao Tian, Shenglong Ye, Zhangwei Gao, Erfei Cui, Wenwen Tong, Kongzhi Hu, Jiapeng Luo, Zheng Ma, et~al. 2024{\natexlab{c}}.
\newblock \href {https://arxiv.org/pdf/2404.16821} {How far are we to gpt-4v? closing the gap to commercial multimodal models with open-source suites}.
\newblock \emph{Science China Information Sciences}, 67(12):220101.

\bibitem[{Cheng et~al.(2024{\natexlab{a}})Cheng, Li, Xu, Zhang, Zhou, and Liu}]{cheng2024vision}
Kanzhi Cheng, Yantao Li, Fangzhi Xu, Jianbing Zhang, Hao Zhou, and Yang Liu. 2024{\natexlab{a}}.
\newblock \href {https://arxiv.org/pdf/2411.00855} {Vision-language models can self-improve reasoning via reflection}.
\newblock \emph{arXiv preprint arXiv:2411.00855}.

\bibitem[{Cheng et~al.(2022)Cheng, Ma, Zong, Zhang, Dai, and Chen}]{cheng2022ads}
Kanzhi Cheng, Zheng Ma, Shi Zong, Jianbing Zhang, Xinyu Dai, and Jiajun Chen. 2022.
\newblock \href {https://arxiv.org/pdf/2308.01143} {Ads-cap: A framework for accurate and diverse stylized captioning with unpaired stylistic corpora}.
\newblock In \emph{CCF International Conference on Natural Language Processing and Chinese Computing}, pages 736--748. Springer.

\bibitem[{Cheng et~al.(2023)Cheng, Song, Ma, Zhu, Zhu, and Zhang}]{cheng2023beyond}
Kanzhi Cheng, Wenpo Song, Zheng Ma, Wenhao Zhu, Zixuan Zhu, and Jianbing Zhang. 2023.
\newblock \href {https://arxiv.org/pdf/2308.01126} {Beyond generic: Enhancing image captioning with real-world knowledge using vision-language pre-training model}.
\newblock In \emph{Proceedings of the 31st ACM International Conference on Multimedia}, pages 5038--5047.

\bibitem[{Cheng et~al.(2024{\natexlab{b}})Cheng, Sun, Chu, Xu, Li, Zhang, and Wu}]{cheng2024seeclick}
Kanzhi Cheng, Qiushi Sun, Yougang Chu, Fangzhi Xu, Yantao Li, Jianbing Zhang, and Zhiyong Wu. 2024{\natexlab{b}}.
\newblock \href {https://arxiv.org/pdf/2401.10935} {Seeclick: Harnessing gui grounding for advanced visual gui agents}.
\newblock \emph{arXiv preprint arXiv:2401.10935}.

\bibitem[{Chiang et~al.(2024)Chiang, Zheng, Sheng, Angelopoulos, Li, Li, Zhang, Zhu, Jordan, Gonzalez et~al.}]{chiang2024chatbot}
Wei-Lin Chiang, Lianmin Zheng, Ying Sheng, Anastasios~Nikolas Angelopoulos, Tianle Li, Dacheng Li, Hao Zhang, Banghua Zhu, Michael Jordan, Joseph~E Gonzalez, et~al. 2024.
\newblock \href {https://arxiv.org/pdf/2403.04132} {Chatbot arena: An open platform for evaluating llms by human preference}.
\newblock \emph{arXiv preprint arXiv:2403.04132}.

\bibitem[{Chou et~al.(2024)Chou, Dunlap, Mashita, Mandal, Darrell, Stoica, Gonzalez, and Chiang}]{chou2024visionarena}
Christopher Chou, Lisa Dunlap, Koki Mashita, Krishna Mandal, Trevor Darrell, Ion Stoica, Joseph~E Gonzalez, and Wei-Lin Chiang. 2024.
\newblock \href {https://arxiv.org/pdf/2412.08687} {Visionarena: 230k real world user-vlm conversations with preference labels}.
\newblock \emph{arXiv preprint arXiv:2412.08687}.

\bibitem[{Cornia et~al.(2020)Cornia, Stefanini, Baraldi, and Cucchiara}]{cornia2020meshed}
Marcella Cornia, Matteo Stefanini, Lorenzo Baraldi, and Rita Cucchiara. 2020.
\newblock \href {http://openaccess.thecvf.com/content_CVPR_2020/papers/Cornia_Meshed-Memory_Transformer_for_Image_Captioning_CVPR_2020_paper.pdf} {Meshed-memory transformer for image captioning}.
\newblock In \emph{Proceedings of the IEEE/CVF conference on computer vision and pattern recognition}, pages 10578--10587.

\bibitem[{Dong et~al.(2024)Dong, Li, Wu, Wang, Zhang, and Guo}]{dong2024benchmarking}
Hongyuan Dong, Jiawen Li, Bohong Wu, Jiacong Wang, Yuan Zhang, and Haoyuan Guo. 2024.
\newblock \href {https://arxiv.org/pdf/2405.19092} {Benchmarking and improving detail image caption}.
\newblock \emph{arXiv preprint arXiv:2405.19092}.

\bibitem[{Hessel et~al.(2021)Hessel, Holtzman, Forbes, Bras, and Choi}]{hessel2021clipscore}
Jack Hessel, Ari Holtzman, Maxwell Forbes, Ronan~Le Bras, and Yejin Choi. 2021.
\newblock \href {https://arxiv.org/pdf/2104.08718} {Clipscore: A reference-free evaluation metric for image captioning}.
\newblock \emph{arXiv preprint arXiv:2104.08718}.

\bibitem[{Hodosh et~al.(2013)Hodosh, Young, and Hockenmaier}]{hodosh2013framing}
Micah Hodosh, Peter Young, and Julia Hockenmaier. 2013.
\newblock \href {https://www.jair.org/index.php/jair/article/download/10833/25854} {Framing image description as a ranking task: Data, models and evaluation metrics}.
\newblock \emph{Journal of Artificial Intelligence Research}, 47:853--899.

\bibitem[{Kasai et~al.(2021)Kasai, Sakaguchi, Dunagan, Morrison, Bras, Choi, and Smith}]{kasai2021transparent}
Jungo Kasai, Keisuke Sakaguchi, Lavinia Dunagan, Jacob Morrison, Ronan~Le Bras, Yejin Choi, and Noah~A Smith. 2021.
\newblock \href {https://arxiv.org/pdf/2111.08940} {Transparent human evaluation for image captioning}.
\newblock \emph{arXiv preprint arXiv:2111.08940}.

\bibitem[{Lee et~al.(2024)Lee, Park, and Kang}]{lee2024fleur}
Yebin Lee, Imseong Park, and Myungjoo Kang. 2024.
\newblock \href {https://arxiv.org/pdf/2406.06004} {Fleur: An explainable reference-free evaluation metric for image captioning using a large multimodal model}.
\newblock \emph{arXiv preprint arXiv:2406.06004}.

\bibitem[{Li et~al.(2024{\natexlab{a}})Li, Zhang, Guo, Zhang, Li, Zhang, Zhang, Zhang, Li, Liu et~al.}]{li2024llava}
Bo~Li, Yuanhan Zhang, Dong Guo, Renrui Zhang, Feng Li, Hao Zhang, Kaichen Zhang, Peiyuan Zhang, Yanwei Li, Ziwei Liu, et~al. 2024{\natexlab{a}}.
\newblock \href {https://arxiv.org/pdf/2408.03326} {Llava-onevision: Easy visual task transfer}.
\newblock \emph{arXiv preprint arXiv:2408.03326}.

\bibitem[{Li et~al.(2024{\natexlab{b}})Li, Wei, Xie, Yang, Song, Wang, An, Liu, Li, Lin et~al.}]{li2024vlrewardbench}
Lei Li, Yuancheng Wei, Zhihui Xie, Xuqing Yang, Yifan Song, Peiyi Wang, Chenxin An, Tianyu Liu, Sujian Li, Bill~Yuchen Lin, et~al. 2024{\natexlab{b}}.
\newblock \href {https://arxiv.org/pdf/2411.17451} {Vlrewardbench: A challenging benchmark for vision-language generative reward models}.
\newblock \emph{arXiv preprint arXiv:2411.17451}.

\bibitem[{Li et~al.(2024{\natexlab{c}})Li, Chiang, Frick, Dunlap, Wu, Zhu, Gonzalez, and Stoica}]{li2024crowdsourced}
Tianle Li, Wei-Lin Chiang, Evan Frick, Lisa Dunlap, Tianhao Wu, Banghua Zhu, Joseph~E Gonzalez, and Ion Stoica. 2024{\natexlab{c}}.
\newblock \href {https://arxiv.org/pdf/2406.11939} {From crowdsourced data to high-quality benchmarks: Arena-hard and benchbuilder pipeline}.
\newblock \emph{arXiv preprint arXiv:2406.11939}.

\bibitem[{Li et~al.(2023)Li, Du, Zhou, Wang, Zhao, and Wen}]{li2023evaluating}
Yifan Li, Yifan Du, Kun Zhou, Jinpeng Wang, Wayne~Xin Zhao, and Ji-Rong Wen. 2023.
\newblock \href {https://arxiv.org/pdf/2305.10355} {Evaluating object hallucination in large vision-language models}.
\newblock \emph{arXiv preprint arXiv:2305.10355}.

\bibitem[{Lin et~al.(2024)Lin, Deng, Chandu, Brahman, Ravichander, Pyatkin, Dziri, Bras, and Choi}]{lin2024wildbench}
Bill~Yuchen Lin, Yuntian Deng, Khyathi Chandu, Faeze Brahman, Abhilasha Ravichander, Valentina Pyatkin, Nouha Dziri, Ronan~Le Bras, and Yejin Choi. 2024.
\newblock \href {https://arxiv.org/pdf/2406.04770} {Wildbench: Benchmarking llms with challenging tasks from real users in the wild}.
\newblock \emph{arXiv preprint arXiv:2406.04770}.

\bibitem[{Lin et~al.(2014)Lin, Maire, Belongie, Hays, Perona, Ramanan, Doll{\'a}r, and Zitnick}]{lin2014microsoft}
Tsung-Yi Lin, Michael Maire, Serge Belongie, James Hays, Pietro Perona, Deva Ramanan, Piotr Doll{\'a}r, and C~Lawrence Zitnick. 2014.
\newblock \href {https://core.ac.uk/download/pdf/216302137.pdf} {Microsoft coco: Common objects in context}.
\newblock In \emph{Computer Vision--ECCV 2014: 13th European Conference, Zurich, Switzerland, September 6-12, 2014, Proceedings, Part V 13}, pages 740--755. Springer.

\bibitem[{Liu et~al.(2024)Liu, Li, Wu, and Lee}]{liu2024visual}
Haotian Liu, Chunyuan Li, Qingyang Wu, and Yong~Jae Lee. 2024.
\newblock \href {https://proceedings.neurips.cc/paper_files/paper/2023/file/6dcf277ea32ce3288914faf369fe6de0-Paper-Conference.pdf} {Visual instruction tuning}.
\newblock \emph{Advances in neural information processing systems}, 36.

\bibitem[{Lu et~al.(2024)Lu, Wu, Zheng, Ma, Gong, Liu, Zhai, Cao, Shen, and Zha}]{lu2024benchmarking}
Fan Lu, Wei Wu, Kecheng Zheng, Shuailei Ma, Biao Gong, Jiawei Liu, Wei Zhai, Yang Cao, Yujun Shen, and Zheng-Jun Zha. 2024.
\newblock \href {https://arxiv.org/pdf/2412.08614} {Benchmarking large vision-language models via directed scene graph for comprehensive image captioning}.
\newblock \emph{arXiv preprint arXiv:2412.08614}.

\bibitem[{Ma et~al.(2023)Ma, Pan, Wu, Cheng, Zhang, Huang, and Chen}]{ma2023food}
Zheng Ma, Mianzhi Pan, Wenhan Wu, Kanzhi Cheng, Jianbing Zhang, Shujian Huang, and Jiajun Chen. 2023.
\newblock \href {https://arxiv.org/pdf/2308.03151} {Food-500 cap: A fine-grained food caption benchmark for evaluating vision-language models}.
\newblock In \emph{Proceedings of the 31st ACM International Conference on Multimedia}, pages 5674--5685.

\bibitem[{Onoe et~al.(2024)Onoe, Rane, Berger, Bitton, Cho, Garg, Ku, Parekh, Pont-Tuset, Tanzer et~al.}]{onoe2024docci}
Yasumasa Onoe, Sunayana Rane, Zachary Berger, Yonatan Bitton, Jaemin Cho, Roopal Garg, Alexander Ku, Zarana Parekh, Jordi Pont-Tuset, Garrett Tanzer, et~al. 2024.
\newblock \href {https://arxiv.org/pdf/2404.19753?} {Docci: Descriptions of connected and contrasting images}.
\newblock In \emph{European Conference on Computer Vision}, pages 291--309. Springer.

\bibitem[{OpenAI(2023)}]{openai2023gpt4}
OpenAI. 2023.
\newblock \href {https://arxiv.org/abs/2303.08774} {{GPT-4} technical report}.
\newblock \emph{Preprint}, arXiv:2303.08774.

\bibitem[{Papineni et~al.(2002)Papineni, Roukos, Ward, and Zhu}]{papineni2002bleu}
Kishore Papineni, Salim Roukos, Todd Ward, and Wei-Jing Zhu. 2002.
\newblock \href {https://aclanthology.org/P02-1040.pdf} {Bleu: a method for automatic evaluation of machine translation}.
\newblock In \emph{Proceedings of the 40th annual meeting of the Association for Computational Linguistics}, pages 311--318.

\bibitem[{Pi et~al.(2024)Pi, Zhang, Zhang, Pan, Chen, and Zhang}]{pi2024image}
Renjie Pi, Jianshu Zhang, Jipeng Zhang, Rui Pan, Zhekai Chen, and Tong Zhang. 2024.
\newblock \href {https://arxiv.org/pdf/2406.07502} {Image textualization: An automatic framework for creating accurate and detailed image descriptions}.
\newblock \emph{arXiv preprint arXiv:2406.07502}.

\bibitem[{Radford et~al.(2021)Radford, Kim, Hallacy, Ramesh, Goh, Agarwal, Sastry, Askell, Mishkin, Clark et~al.}]{radford2021learning}
Alec Radford, Jong~Wook Kim, Chris Hallacy, Aditya Ramesh, Gabriel Goh, Sandhini Agarwal, Girish Sastry, Amanda Askell, Pamela Mishkin, Jack Clark, et~al. 2021.
\newblock \href {http://proceedings.mlr.press/v139/radford21a/radford21a.pdf} {Learning transferable visual models from natural language supervision}.
\newblock In \emph{International conference on machine learning}, pages 8748--8763. PMLR.

\bibitem[{Sun et~al.(2024)Sun, Cheng, Ding, Jin, Wang, Xu, Wu, Jia, Chen, Liu et~al.}]{sun2024genesis}
Qiushi Sun, Kanzhi Cheng, Zichen Ding, Chuanyang Jin, Yian Wang, Fangzhi Xu, Zhenyu Wu, Chengyou Jia, Liheng Chen, Zhoumianze Liu, et~al. 2024.
\newblock \href {https://arxiv.org/pdf/2412.19723} {Os-genesis: Automating gui agent trajectory construction via reverse task synthesis}.
\newblock \emph{arXiv preprint arXiv:2412.19723}.

\bibitem[{Touvron et~al.(2023)Touvron, Lavril, Izacard, Martinet, Lachaux, Lacroix, Rozi{\`e}re, Goyal, Hambro, Azhar et~al.}]{touvron2023llama}
Hugo Touvron, Thibaut Lavril, Gautier Izacard, Xavier Martinet, Marie-Anne Lachaux, Timoth{\'e}e Lacroix, Baptiste Rozi{\`e}re, Naman Goyal, Eric Hambro, Faisal Azhar, et~al. 2023.
\newblock \href {https://arxiv.org/pdf/2302.13971} {Llama: Open and efficient foundation language models}.
\newblock \emph{arXiv preprint arXiv:2302.13971}.

\bibitem[{Vedantam et~al.(2015)Vedantam, Lawrence~Zitnick, and Parikh}]{vedantam2015cider}
Ramakrishna Vedantam, C~Lawrence~Zitnick, and Devi Parikh. 2015.
\newblock \href {https://openaccess.thecvf.com/content_cvpr_2015/papers/Vedantam_CIDEr_Consensus-Based_Image_2015_CVPR_paper.pdf} {Cider: Consensus-based image description evaluation}.
\newblock In \emph{Proceedings of the IEEE conference on computer vision and pattern recognition}, pages 4566--4575.

\bibitem[{Vinyals et~al.(2015)Vinyals, Toshev, Bengio, and Erhan}]{vinyals2015show}
Oriol Vinyals, Alexander Toshev, Samy Bengio, and Dumitru Erhan. 2015.
\newblock \href {http://openaccess.thecvf.com/content_cvpr_2015/papers/Vinyals_Show_and_Tell_2015_CVPR_paper.pdf} {Show and tell: A neural image caption generator}.
\newblock In \emph{Proceedings of the IEEE conference on computer vision and pattern recognition}, pages 3156--3164.

\bibitem[{Wada et~al.(2024)Wada, Kaneda, Saito, and Sugiura}]{wada2024polos}
Yuiga Wada, Kanta Kaneda, Daichi Saito, and Komei Sugiura. 2024.
\newblock \href {https://openaccess.thecvf.com/content/CVPR2024/papers/Wada_Polos_Multimodal_Metric_Learning_from_Human_Feedback_for_Image_Captioning_CVPR_2024_paper.pdf} {Polos: Multimodal metric learning from human feedback for image captioning}.
\newblock In \emph{Proceedings of the IEEE/CVF Conference on Computer Vision and Pattern Recognition}, pages 13559--13568.

\bibitem[{Wang et~al.(2022)Wang, Yang, Men, Lin, Bai, Li, Ma, Zhou, Zhou, and Yang}]{wang2022ofa}
Peng Wang, An~Yang, Rui Men, Junyang Lin, Shuai Bai, Zhikang Li, Jianxin Ma, Chang Zhou, Jingren Zhou, and Hongxia Yang. 2022.
\newblock \href {https://arxiv.org/pdf/2202.03052} {Ofa: Unifying architectures, tasks, and modalities through a simple sequence-to-sequence learning framework}.
\newblock In \emph{International conference on machine learning}, pages 23318--23340. PMLR.

\bibitem[{Wu et~al.(2024)Wu, Wu, Xu, Wang, Sun, Jia, Cheng, Ding, Chen, Liang et~al.}]{wu2024atlas}
Zhiyong Wu, Zhenyu Wu, Fangzhi Xu, Yian Wang, Qiushi Sun, Chengyou Jia, Kanzhi Cheng, Zichen Ding, Liheng Chen, Paul~Pu Liang, et~al. 2024.
\newblock \href {https://arxiv.org/pdf/2410.23218} {Os-atlas: A foundation action model for generalist gui agents}.
\newblock \emph{arXiv preprint arXiv:2410.23218}.

\bibitem[{Xiong et~al.(2024)Xiong, Wang, Guo, Ye, Fan, Gu, Huang, and Li}]{xiong2024llava}
Tianyi Xiong, Xiyao Wang, Dong Guo, Qinghao Ye, Haoqi Fan, Quanquan Gu, Heng Huang, and Chunyuan Li. 2024.
\newblock \href {https://arxiv.org/pdf/2410.02712} {Llava-critic: Learning to evaluate multimodal models}.
\newblock \emph{arXiv preprint arXiv:2410.02712}.

\bibitem[{Xu(2015)}]{xu2015show}
Kelvin Xu. 2015.
\newblock \href {http://dl1.icdst.org/pdfs/files/32b84efede46be9a8bb0014324bfcdb6.pdf} {Show, attend and tell: Neural image caption generation with visual attention}.
\newblock \emph{arXiv preprint arXiv:1502.03044}.

\bibitem[{Ye et~al.(2023)Ye, Xu, Xu, Ye, Yan, Zhou, Wang, Hu, Shi, Shi et~al.}]{ye2023mplug}
Qinghao Ye, Haiyang Xu, Guohai Xu, Jiabo Ye, Ming Yan, Yiyang Zhou, Junyang Wang, Anwen Hu, Pengcheng Shi, Yaya Shi, et~al. 2023.
\newblock \href {https://arxiv.org/pdf/2304.14178} {mplug-owl: Modularization empowers large language models with multimodality}.
\newblock \emph{arXiv preprint arXiv:2304.14178}.

\bibitem[{Yue et~al.(2024)Yue, Ni, Zhang, Zheng, Liu, Zhang, Stevens, Jiang, Ren, Sun et~al.}]{yue2024mmmu}
Xiang Yue, Yuansheng Ni, Kai Zhang, Tianyu Zheng, Ruoqi Liu, Ge~Zhang, Samuel Stevens, Dongfu Jiang, Weiming Ren, Yuxuan Sun, et~al. 2024.
\newblock \href {https://openaccess.thecvf.com/content/CVPR2024/papers/Yue_MMMU_A_Massive_Multi-discipline_Multimodal_Understanding_and_Reasoning_Benchmark_for_CVPR_2024_paper.pdf} {Mmmu: A massive multi-discipline multimodal understanding and reasoning benchmark for expert agi}.
\newblock In \emph{Proceedings of the IEEE/CVF Conference on Computer Vision and Pattern Recognition}, pages 9556--9567.

\bibitem[{Zhai et~al.(2023)Zhai, Mustafa, Kolesnikov, and Beyer}]{zhai2023sigmoid}
Xiaohua Zhai, Basil Mustafa, Alexander Kolesnikov, and Lucas Beyer. 2023.
\newblock \href {http://openaccess.thecvf.com/content/ICCV2023/papers/Zhai_Sigmoid_Loss_for_Language_Image_Pre-Training_ICCV_2023_paper.pdf} {Sigmoid loss for language image pre-training}.
\newblock In \emph{Proceedings of the IEEE/CVF International Conference on Computer Vision}, pages 11975--11986.

\bibitem[{Zhang et~al.(2024)Zhang, Zhang, Dong, Zang, and Wang}]{zhang2024long}
Beichen Zhang, Pan Zhang, Xiaoyi Dong, Yuhang Zang, and Jiaqi Wang. 2024.
\newblock \href {https://arxiv.org/pdf/2403.15378} {Long-clip: Unlocking the long-text capability of clip}.
\newblock In \emph{European Conference on Computer Vision}, pages 310--325. Springer.

\bibitem[{Zheng et~al.(2023)Zheng, Chiang, Sheng, Zhuang, Wu, Zhuang, Lin, Li, Li, Xing, Zhang, Gonzalez, and Stoica}]{zheng2023judging}
Lianmin Zheng, Wei-Lin Chiang, Ying Sheng, Siyuan Zhuang, Zhanghao Wu, Yonghao Zhuang, Zi~Lin, Zhuohan Li, Dacheng Li, Eric.~P Xing, Hao Zhang, Joseph~E. Gonzalez, and Ion Stoica. 2023.
\newblock \href {https://arxiv.org/abs/2306.05685} {Judging llm-as-a-judge with mt-bench and chatbot arena}.
\newblock \emph{Preprint}, arXiv:2306.05685.

\end{thebibliography}

\appendix

\section{DOCCI Dataset}
\label{app:docci}
The DOCCI dataset consists of long, human-annotated English descriptions for images. These images were collected, curated, and donated for research purposes. Most images in DOCCI are natural scenes captured in both indoor and outdoor settings. The researcher aimed to capture key challenges, such as spatial relations, counting, text rendering, and world knowledge, among others. DOCCI captions require detailed, accurate descriptions of objects, attributes, and actions, with clear and grammatically correct language. The annotations ensure high quality and consistency for evaluating image captioning models. The dataset's diverse scenes and high-quality human-annotated captions make it ideal for our benchmark.

\section{Prompts for generating detailed caption}
\label{app:prompt}

\begin{figure*}[ht]
\centering
\includegraphics[width=1.0\textwidth]{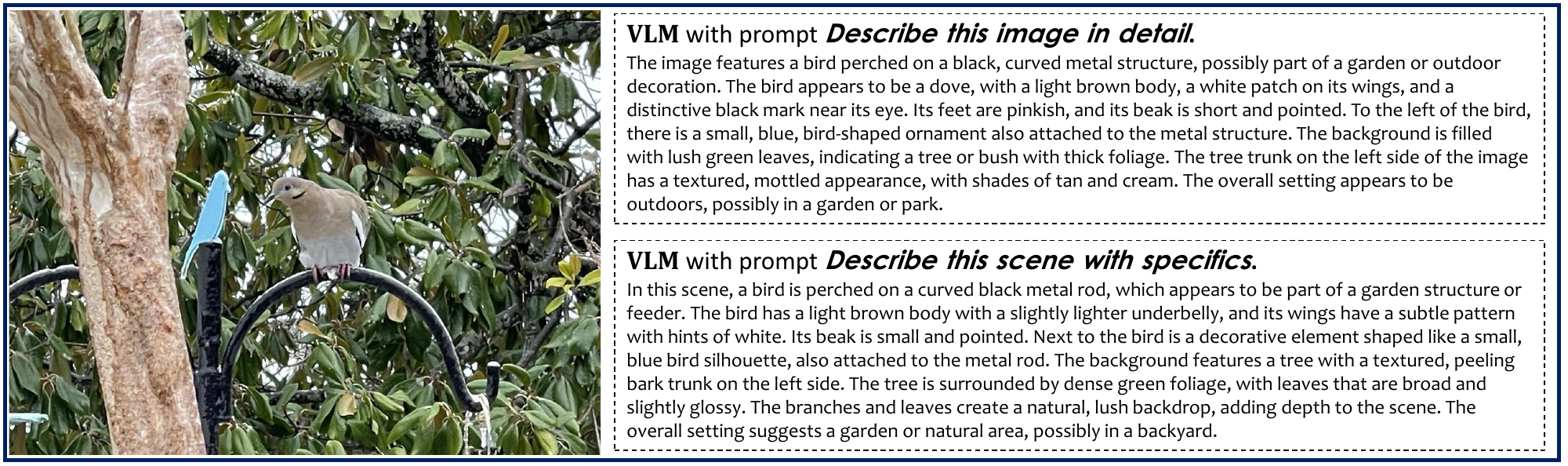}
\caption{Detailed caption generated with different prompt.}
\label{fig:10prompt}
\vspace{-1.0em}
\end{figure*}

\Cref{tab:prompts} provides the list of prompts used to generate detailed captions for images.
Figure \ref{fig:10prompt} shows captions generated with different prompts by the same VLM.

\begin{table}[ht]
\small
\centering
\begin{tabular}{p{0.95\columnwidth}}
\toprule
\textbf{Prompts for Caption Generating} \\
\midrule
Describe this image in detail. \\
Generate a detailed caption for this image. \\
Explain the contents of this photo with specifics. \\
Create a rich and descriptive caption for this visual content. \\
Give this image a detailed image caption. \\
Write a detailed explanation of the scene depicted here. \\
Provide a detailed description of this photo. \\
Provide a thorough description of this image. \\
Describe this scene with specifics. \\
Provide a descriptive caption for this image. \\
\bottomrule
\end{tabular}
\caption{Prompts for detailed captioning.}
\label{tab:prompts}
\end{table}

\section{Model List uesd in annotation}
\label{app:model}
The models selected for our experiments include both commercial and open-source models. Among the commercial models, we utilize GPT-4o-0806, GPT-4o-mini-0718, Gemini-2.0-flash-exp, Gemini-1.5-pro-002, and Claude-3.5-Sonnet-0620. On the other hand, we also incorporate several open-source models, including LLama-3.2-90B-Vision-Instruct, Qwen2-VL-72B-Instruct, Qwen2-VL-7B-Instruct, Qwen2-VL-2B-Instruct, LLaVA-v1.6-34B, LLaVA-v1.5-7B, InternVL2-26B, CogVLM2-llama3-chat-19B, and MiniCPM-V2.6-8B. By evaluating both commercial and open-source models, we aim to provide a comprehensive comparison of VLMs in the context of detailed captioning.

\section{GPT-4o vs Human Example}
We observed that human captions occasionally overlook certain image details. For instance, in the upper image of Table \ref{tab:GPTvsHuman}, compared to humans, GPT-4o paid attention to the surrounding environment in greater detail. Additionally, its mention of "a red leash is attached to its collar, trailing across the grass" was more precise than simply stating that "a red leash leading to the bottom of the image". In the lower image, GPT-4o provided a more detailed description of the yacht and also noticed the poster on the window, a feature that was not mentioned by the human annotators.

\label{app:GPTvsHuman}
\begin{table*}[ht]
\centering
\small
\resizebox{1.0\linewidth}{!}{%
\begin{tabular}{p{0.37\textwidth}p{0.63\textwidth}}
\toprule
\centering
\textbf{Image} & \textbf{Detailed Caption}\\
\midrule
 & \\
\begin{minipage}[c]{.37\textwidth}
\centering

\includegraphics[height=38mm]{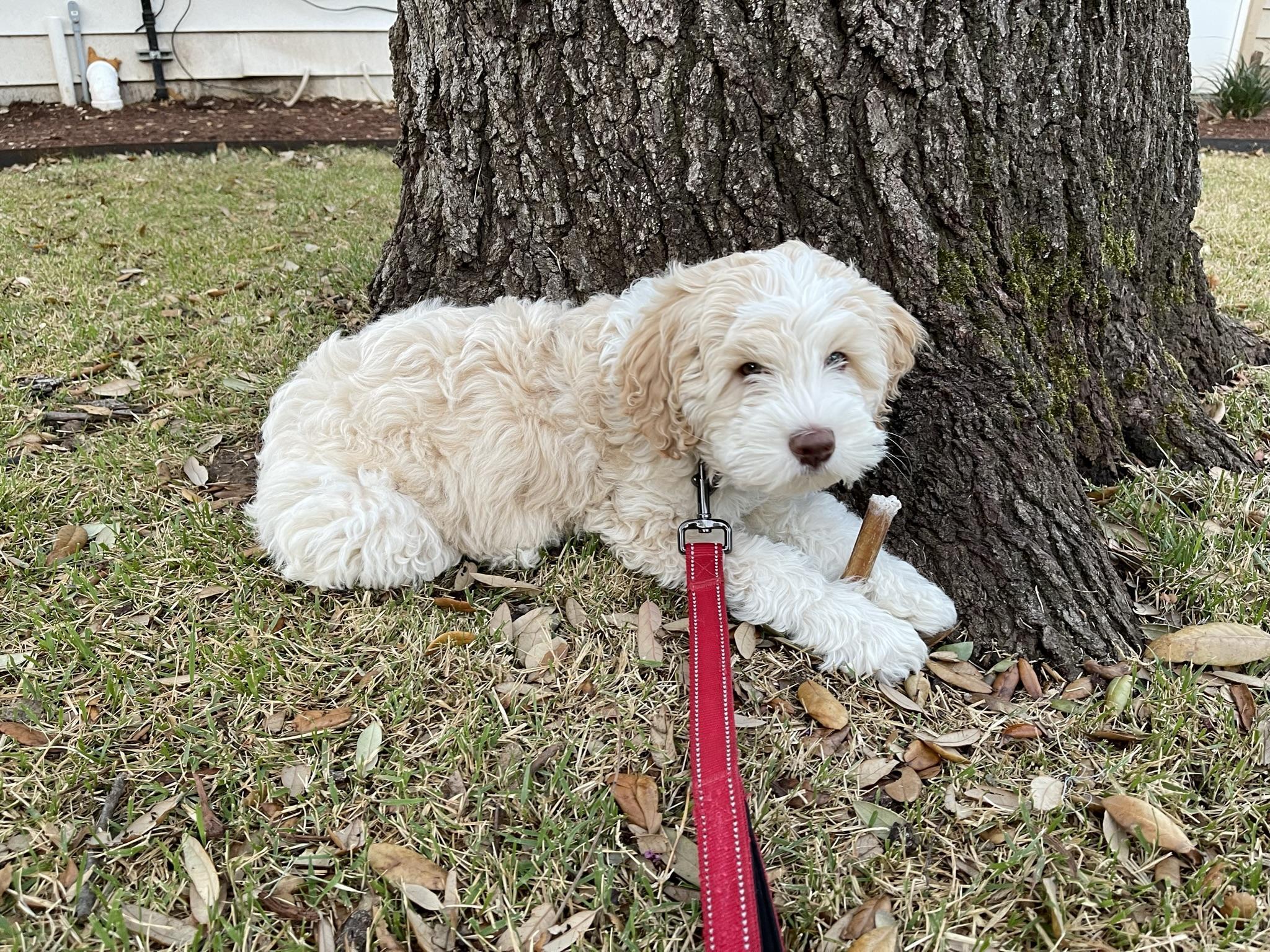}
\end{minipage}
& 
\begin{minipage}[c]{.63\textwidth}
\textbf{human: }A dog laying on it's belly looking at the camera, The dog appears to be by a tree on a lawn of grass with dead leaves around the area. The dog has white and curly fur, It has a bone held by its paws, and has a red leash leading to the bottom of the image. The dog is looking directly at the camera and has its body shifted to the right.\\
\textbf{GPT-4o-0806}: A fluffy, cream-colored puppy is resting against the trunk of a large tree in a grassy area. The puppy has a soft, curly coat and is holding a chew stick in its front paws. \greenmark{A red leash is attached to its collar}, trailing across the grass. The ground is scattered with dry leaves, \greenmark{and the background shows part} \greenmark{of a house with some garden elements}. The puppy looks content and relaxed in its outdoor setting.

\end{minipage}

\\
\midrule
 & \\
\begin{minipage}[c]{.37\textwidth}
\centering

\includegraphics[height=40mm]{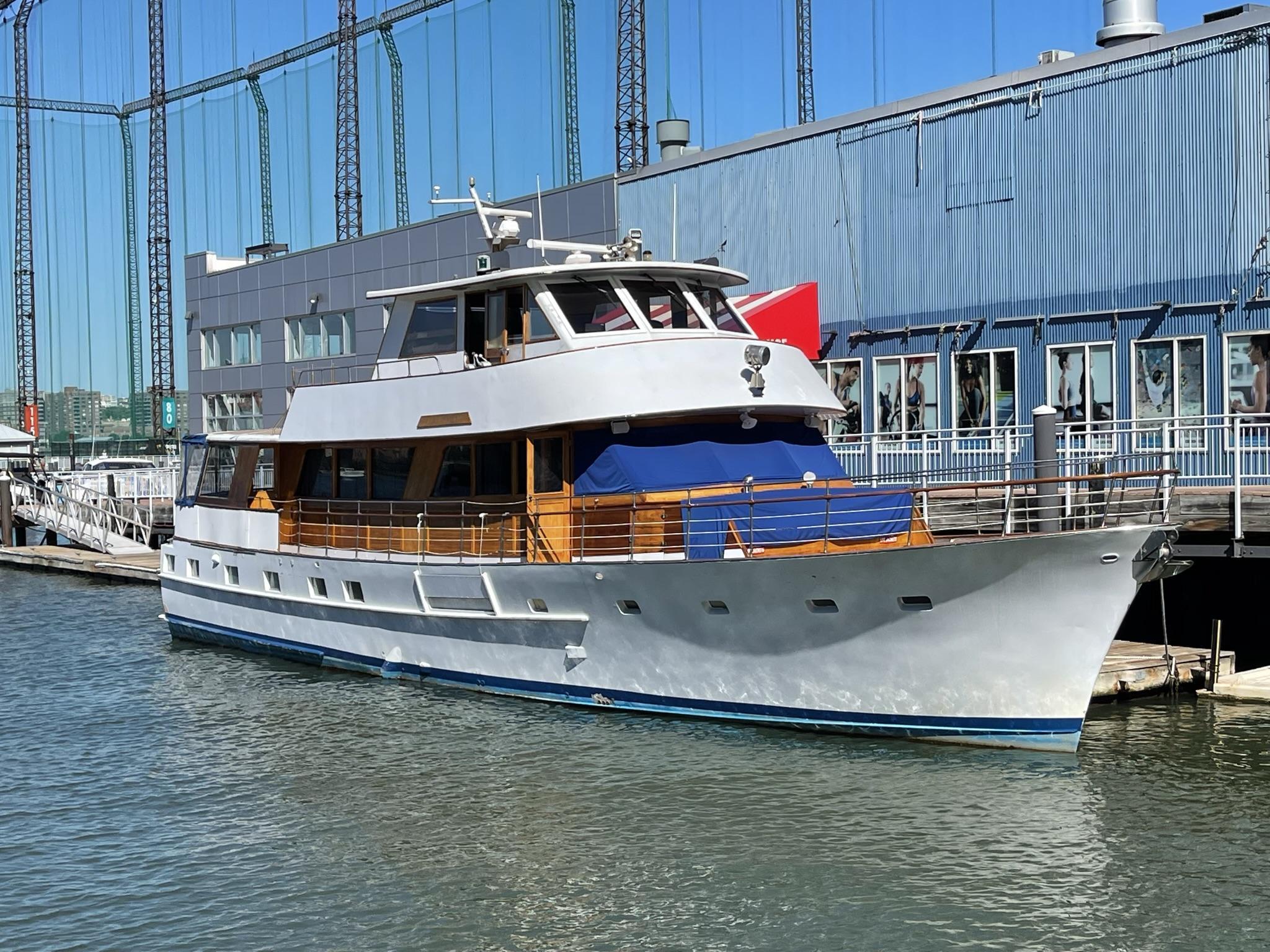}
\end{minipage}
& 
\begin{minipage}[c]{.63\textwidth}
 \textbf{human: }A medium three quarters frontal view of a white yacht on a dock near a body of water that is partially visible at the bottom of the view. Behind the yacht is a large blue building with a blue toned color on the right side of the view, and on its left side the color is gray. The building extends from the middle left of the view to the right side. At the top of the view the blue sky is visible behind large metal constructs and poles that stand above the yacht and nearby building. Between the metal columns and poles are large nets that extend across the upper portion of the view. Reflections of the white yahct are visible on the water at the bottom of the view. The yacht has blue accents around the bottom and has two different deck sections.\\
\textbf{GPT-4o-0806}: The image depicts a large yacht docked at a marina. \greenmark{The yacht} \greenmark{has a sleek, modern design with a white hull and a blue stripe near the waterline}. \greenmark{The upper structure features a combination of white and wooden elements, giv-}\\\greenmark{ing it a classic and elegant appearance}. The main deck has wooden railings and a covered area with blue fabric, likely for protection against the elements. The yacht has multiple levels, with large windows on the lower deck and a spacious upper deck area, possibly for navigation and leisure. There are antennas and equipment on the top, indicating communication or navigation systems. In the background, \greenmark{there is a large building with a blue exterior and multiple windows}, \greenmark{some of which display images or posters}. The building is adjacent to the marina, with a walkway and railings visible alongside the water. Above the building, there are tall metal structures, possibly part of a nearby facility or infrastructure. The water is calm, reflecting parts of the yacht and the surrounding structures. \greenmark{The sky is clear, suggesting a sunny day}.

\end{minipage}

\\

\bottomrule
\end{tabular}
}
\caption{Examples of GPT-4o perform better than human}.
\label{tab:GPTvsHuman}
\end{table*}

\section{Quality Control}
\label{app:qualitycontrol}
Throughout the annotation process, experts intermittently check the annotations of each annotator to ensure consistent high-quality results. When necessary, specific feedback was provided to guide the annotator and enhance the accuracy and consistency of subsequent annotations.

Additionally, the platform continuously monitored the annotators' activity. If an annotator’s annotation time consistently fell below the average, the platform would trigger a feedback alert, prompting experts to pay closer attention to the quality of that annotator's work. Moreover, given the task's complexity, submissions made in under five seconds were considered as potential malicious attempts and were disregarded to maintain the integrity of the data. The rigorous annotator training procedures and stringent platform filtering mechanisms implemented throughout the annotation process ensured the collection of high-quality data.

\section{Protocol Example}
\label{app:protocol-example}
We have included additional annotation examples in Table \ref{tab:full-protocol-example} to offer a clearer and more detailed illustration of our evaluation protocol.
In the annotation process, annotators follow the established guidelines to carefully assess whether any issues exist within the pair of captions for a given image. They compare the two captions against each other, considering multiple aspects. Based on this comparison, annotators provide a annotation result, ensuring that the final result reflects the strengths and weaknesses of each caption. If it is difficult to distinguish between the captions, annotators are also allowed to mark the result as tie.

\begin{table*}[ht]
\centering
\small
\resizebox{1.0\linewidth}{!}{%
\begin{tabular}{p{0.15\textwidth}p{0.7\textwidth}p{0.15\textwidth}}
\toprule
\textbf{Image} & \textbf{Pairwise Caption Battle} & \textbf{Preference} \\
\midrule
\begin{minipage}{.15\textwidth}
\includegraphics[height=20mm]{latex/images/examples/test_00727.jpg}
\end{minipage}
& 
\begin{minipage}{.7\textwidth}
\textbf{Caption1 (Qwen2-VL-7B):} ...  The background shows more cars parked outside, \redmark{indicating that the dealership has a large inventory}. The showroom also features a high ceiling with large windows, allowing for ample natural light.  ... 
\\
\textbf{Caption2 (Qwen2-VL-72B):} ... Outside the large windows, the view shows \greenmark{a parking} \greenmark{lot with several cars parked, including a mix of sedans and SUVs}. The parking lot is surrounded by greenery, with trees and bushes visible in the distance. The sky is partly cloudy, with patches of blue sky and white clouds. ... 
\end{minipage}
& 
\begin{minipage}{.15\textwidth}
    Caption2 is better (Caption1 has baseless speculation).
\end{minipage}
\\
\midrule
\begin{minipage}{.15\textwidth}
\includegraphics[height=20mm]{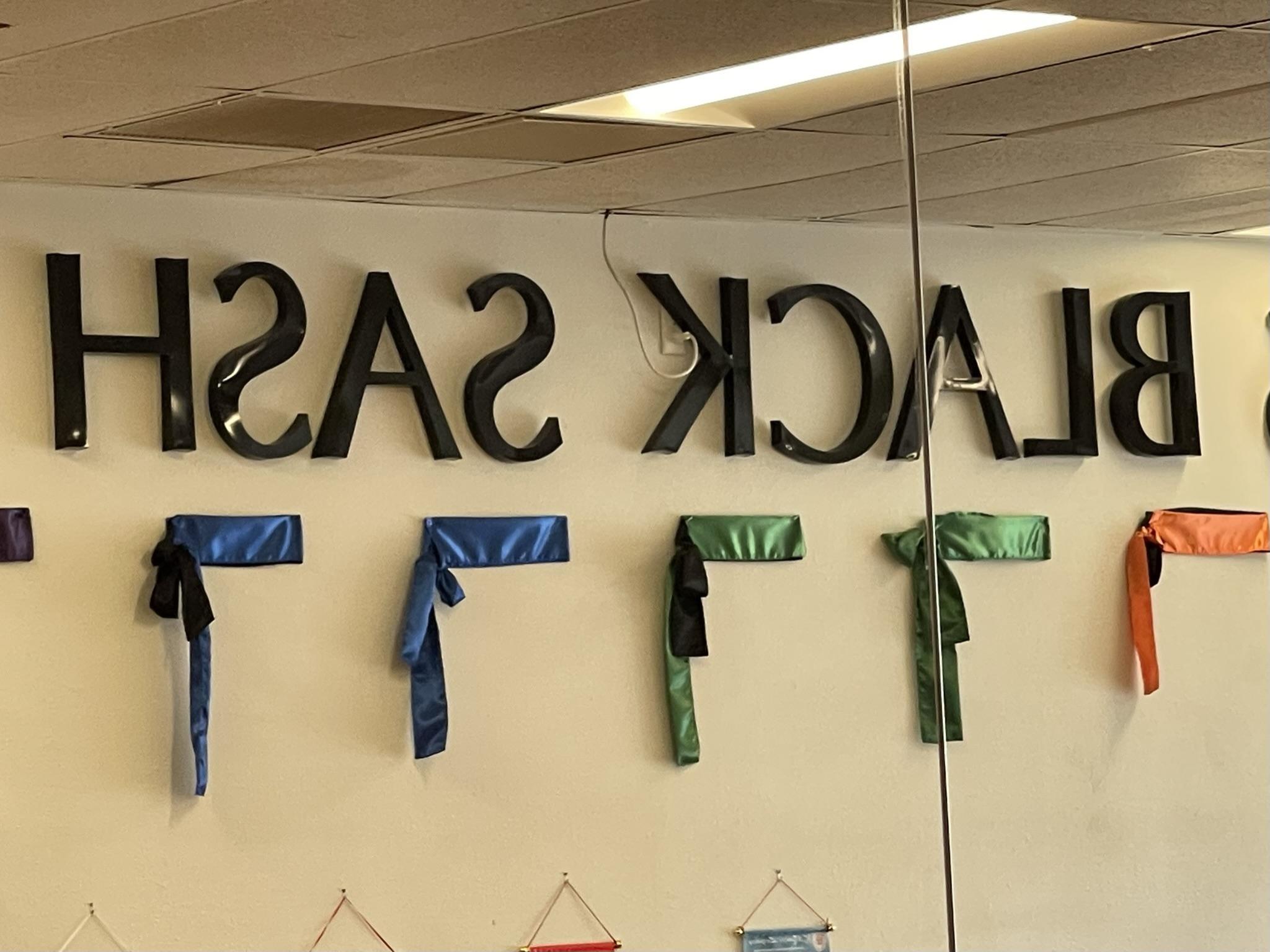}
\end{minipage}
& 
\begin{minipage}{.7\textwidth}
\textbf{Caption1 (Gemini-2.0-flash-exp):} ... The most prominent feature is a large, \greenmark{reversed text} \greenmark{that reads "BLACK SASH"} when viewed correctly (likely through a window). The letters are black, three-dimensional, and seemingly made of a hard, glossy material. ... 
\\
\textbf{Caption2 (MiniCPM-V2.6-8B):} The image shows a wall with the word \redmark{"BLACK ASH"} written in large, black, capital letters. Below the text, there are several colorful ribbons or banners attached to the wall. ... 
\end{minipage}
& 
\begin{minipage}{.15\textwidth}
    Caption1 is better (correctly point out "reversed" and accurately identify it).
\end{minipage}
\\
\midrule
\begin{minipage}{.15\textwidth}
\includegraphics[height=35mm]{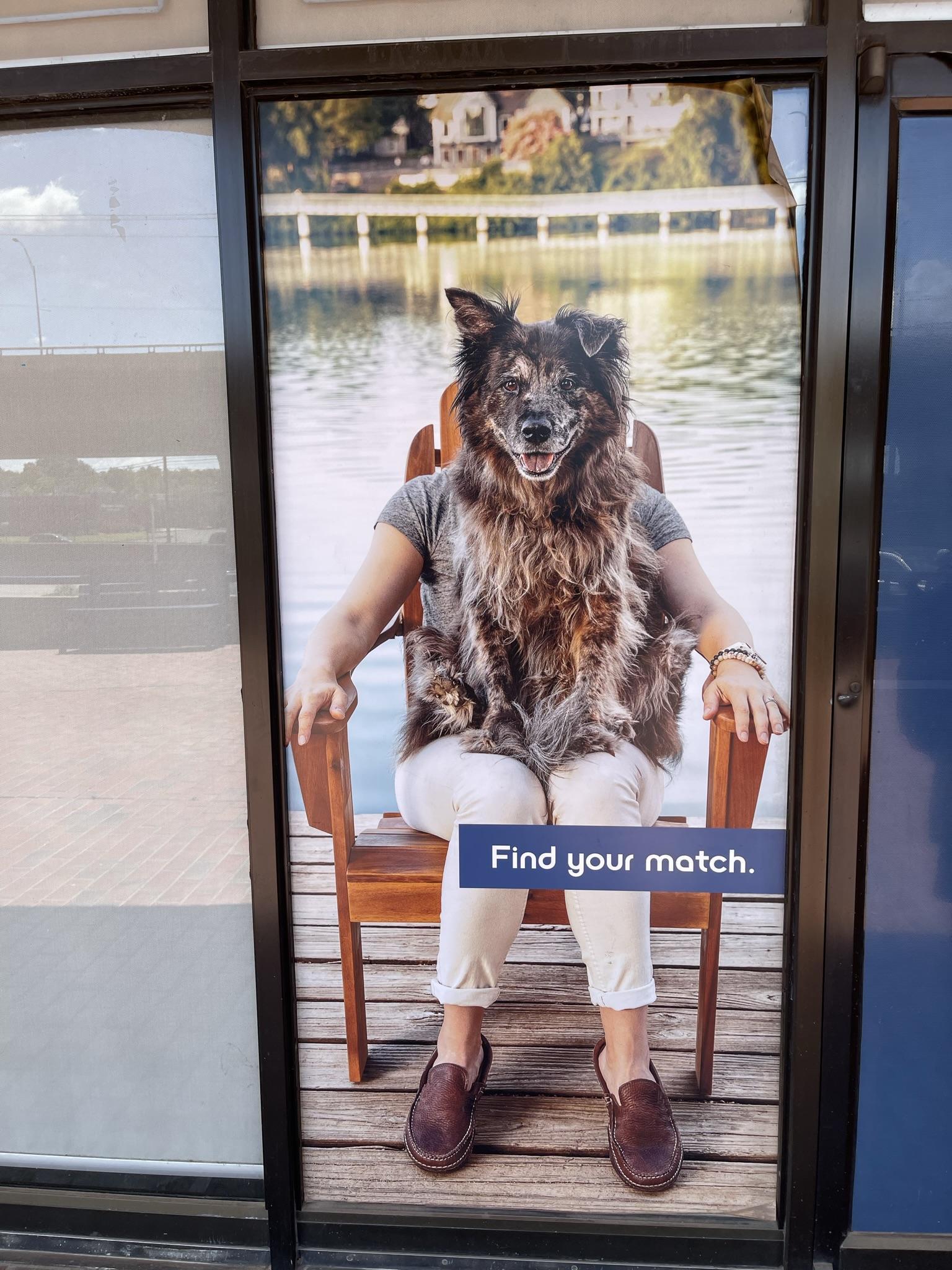}
\end{minipage}
& 
\begin{minipage}{.7\textwidth}
\textbf{Caption1 (GPT-4o-0806):} ... It shows a person sitting on a wooden chair by a lake, \redmark{but instead of a human head, the person has the head of a dog. The dog's head is large} \redmark{and fluffy, with a friendly expression and perked ears}. The person is wearing a gray shirt, white pants, and brown slip-on shoes, and their arms are resting on the armrests of the chair. ... 
\\
\textbf{Caption2 (Claude-3.5-Sonnet):} ...  The scene shows a large, \greenmark{fluffy dog sitting on} \greenmark{someone's lap} in a wooden chair on what looks like a dock or pier by a body of water. The dog, which seems to be a mixed breed with long fur, is front and center, \greenmark{taking up} \greenmark{most of the person's torso and obscuring their upper body}.\ ... 
\end{minipage}
& 
\begin{minipage}{.15\textwidth}
    Caption2 is better (correctly identify the spatial relationship).
\end{minipage}
\\
\midrule
\begin{minipage}{.15\textwidth}
\includegraphics[height=20mm]{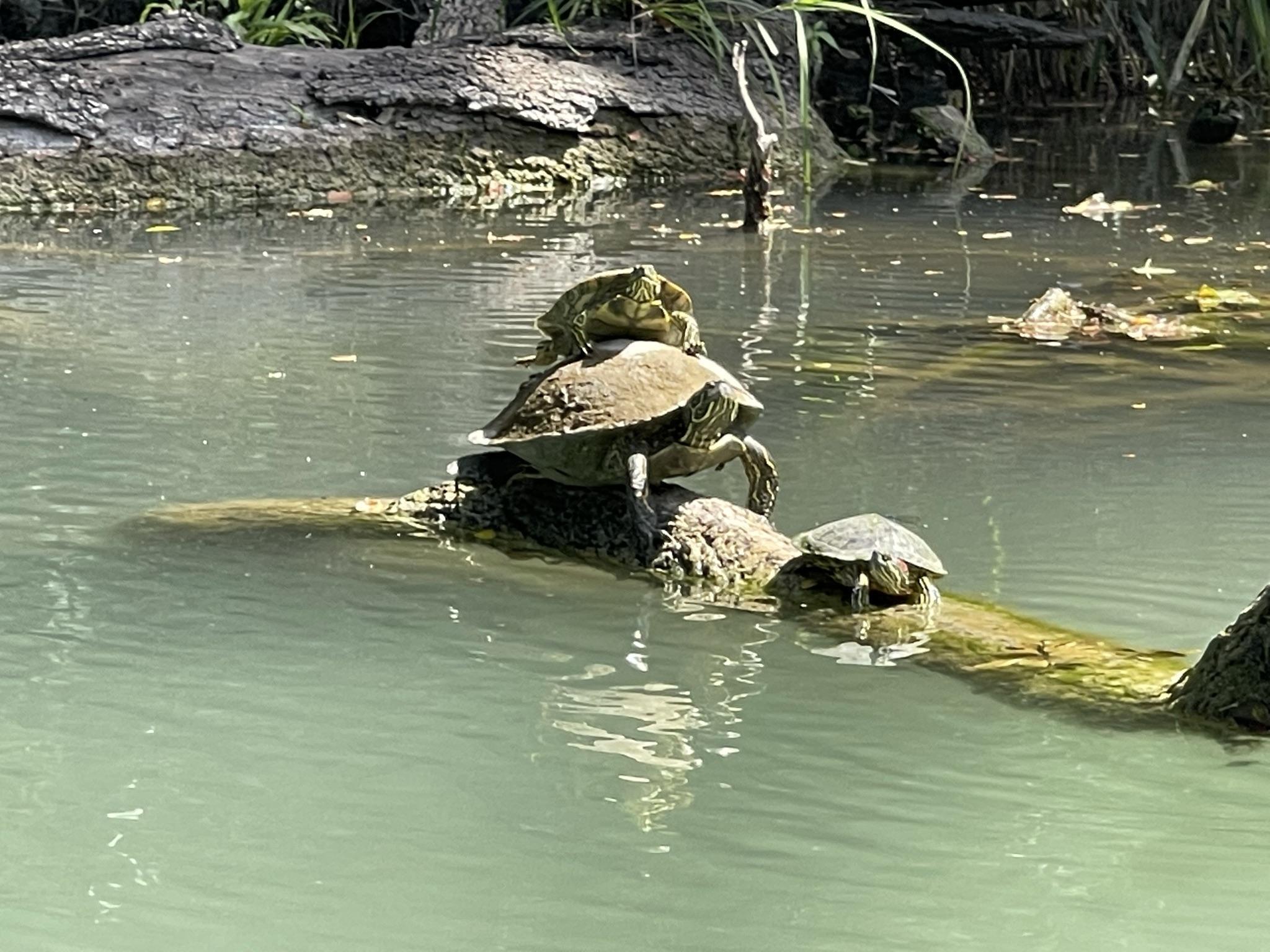}
\end{minipage}
& 
\begin{minipage}{.7\textwidth}
\textbf{Caption1 (Claude-3.5-Sonnet):} ... There are at least three turtles visible, with \redmark{one larger turtle perched on top} and \redmark{two smaller ones below}. The turtles' shells have distinct patterns and colors, ranging from dark green to brown.  ... 
\\
\textbf{Caption2 (human):} ... There is \greenmark{one big turtle and a little small one on top of its shell} and in front of the big turtle there is \greenmark{a small one on the log looking up}.   ... 
\end{minipage}
& 
\begin{minipage}{.15\textwidth}
    Caption2 is better (accurately identify the characteristics of the three turtles).
\end{minipage}
\\
\midrule
\begin{minipage}{.15\textwidth}
\includegraphics[height=20mm]{latex/images/examples/test_04963.jpg}
\end{minipage}
& 
\begin{minipage}{.7\textwidth}
\textbf{Caption1 (Gemini-1.5-pro-002):} A light green and purple monster truck toy is seen in motion, \greenmark{just about to drive off the bottom of a bright green plastic slide}. The toy has large black tires, a pink chassis, and a textured, mottled design on its body that incorporates darker green and black markings, resembling scales or flames. ... 
\\
\textbf{Caption2 (Qwen2-VL-72B):}...The truck is \redmark{positioned on a grassy surface}, and behind it is a large, green, plastic slide. The slide has a smooth, curved surface and is likely part of a playground set. ... 
\end{minipage}
& 
\begin{minipage}{.15\textwidth}
    Caption1 is better (precisely capture the motion state).
\end{minipage}
\\
\midrule
\begin{minipage}{.15\textwidth}
\includegraphics[height=35mm]{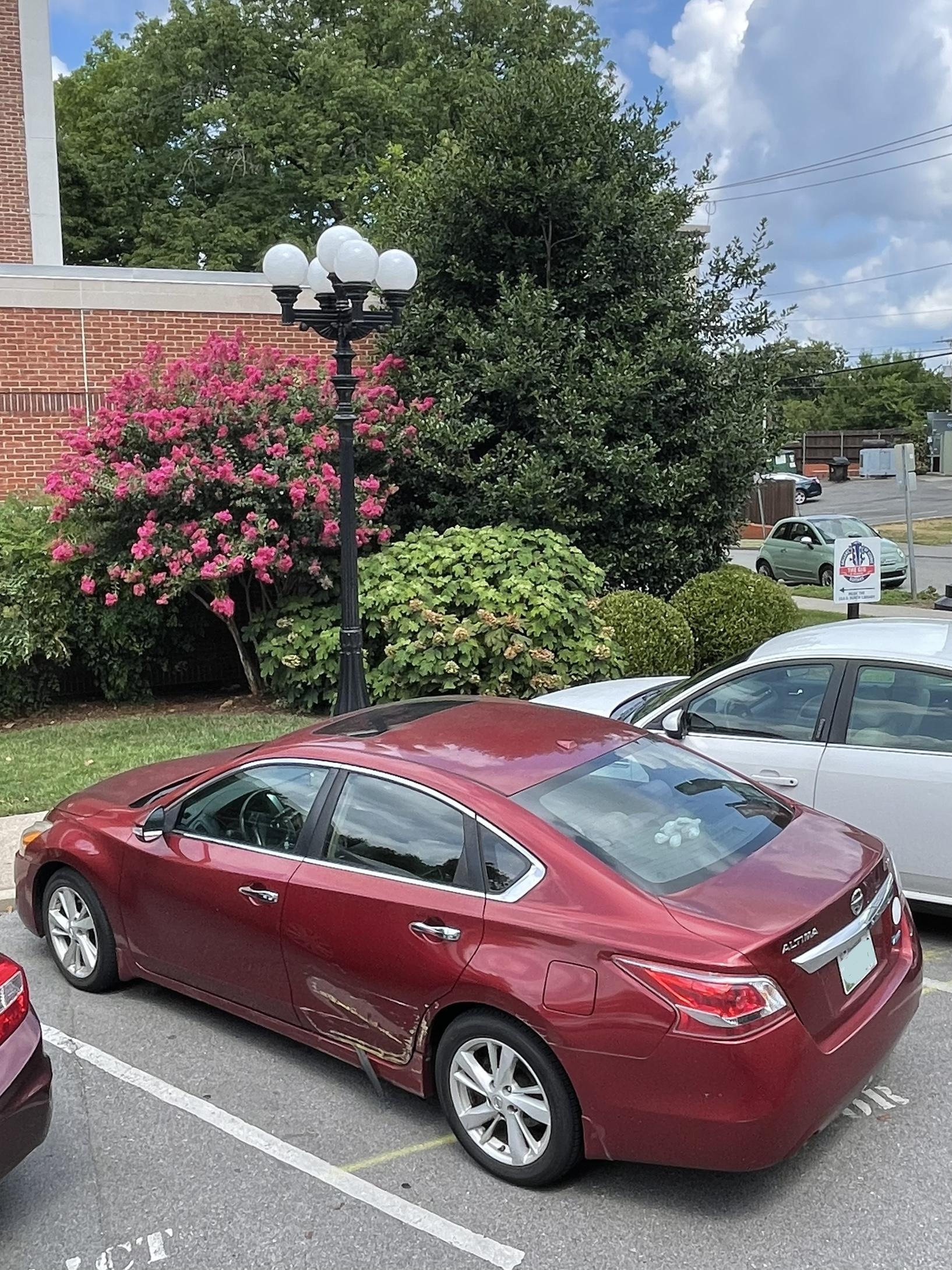}
\end{minipage}
& 
\begin{minipage}{.7\textwidth}
\textbf{Caption1 (GPT-4o-mini-0718):} ... The car is notable for \greenmark{its body damage—specifically}, \greenmark{a significant scrape or dent running along the lower part of its driver’s side}, suggesting that it may have been in a minor collision or scraped against a post or another vehicle.  ... 
\\
\textbf{Caption2 (InternVL2-26B):} ... The car is positioned with its rear facing the camera, and the license plate is visible. \redmark{The car appears to be in good condition, with no apparent} \redmark{damage}. ... 
\end{minipage}
& 
\begin{minipage}{.15\textwidth}
    Caption1 is better (correctly describe the condition of the car).
\end{minipage}
\\
\midrule
\begin{minipage}{.15\textwidth}
\includegraphics[height=35mm]{latex/images/examples/test_00038.jpg}
\end{minipage}
& 
\begin{minipage}{.7\textwidth}
\textbf{Caption1 (human):} ... The shadow depicts a person standing up straight and \greenmark{suspended} \greenmark{in the air while jumping}. The person can be seen wearing shorts, and the edge of a backpack can be seen. ...
\\
\textbf{Caption2 (Gemini-2.0-flash-exp):} ...Shadow Details:...**Pose:** The arms are positioned slightly away from the body and the legs are separated at the feet. \redmark{This indicates a} \redmark{casual, natural standing position}. ...
\end{minipage}
& 
\begin{minipage}{.15\textwidth}
    Caption1 is better (accurately describe the jumping state).
\end{minipage}
\\
\midrule

\begin{minipage}{.15\textwidth}
\includegraphics[height=35mm]{latex/images/examples/test_02693.jpg}
\end{minipage}
& 
\begin{minipage}{.7\textwidth}
\textbf{Caption1 (CogVLM2-llama3-chat-19B):} This image depicts \greenmark{a pink toy car with a} \greenmark{yellow roof}, parked on a \greenmark{paved surface}. The toy car has a simplistic design, \greenmark{with a} \greenmark{steering wheel on the left side} and a single seat. \greenmark{Atop the toy car is a blue helmet}, which is not typically associated with toys. The background shows \greenmark{a wooden fence and a gravel} \greenmark{ground}, suggesting an outdoor setting, possibly a backyard or a park.
\\
\textbf{Caption2 (MiniCPM-V2.6-8B):} The image shows \greenmark{a child's toy car, which is pink} \greenmark{with a beige roof}. The car is \greenmark{equipped with a steering wheel and has a blue helmet} \greenmark{placed on the roof}. The car is situated on \greenmark{a paved area with a wooden fence and gravel} in the background. The toy car appears to be designed for young children, providing a safe and playful environment for imaginative play.
\end{minipage}
& 
\begin{minipage}{.15\textwidth}
    Tie (Both captions accurately describe the appearance of the toy car, the surrounding objects and environment).
\end{minipage}
\\
\bottomrule
\end{tabular}
}
\caption{More example of the evaluation protocol.}
\label{tab:full-protocol-example}
\end{table*}

\section{Failure Cases}
\label{app:failure-cases}

In our analysis of VLM performance, we identified several common mistakes that persist even in models with strong overall performance. We provide four types of representative examples in Table \ref{fig:failure-case}. The last row of the table illustrates typical errors made by weaker models.

\begin{table*}[ht]
\centering
\small
\resizebox{1.0\linewidth}{!}{%
\begin{tabular}{p{0.1\textwidth}p{0.27\textwidth}p{0.63\textwidth}}
\toprule
\textbf{Category} & \textbf{Image} & \textbf{Failure Caption}\\
\midrule

\begin{minipage}{.1\textwidth}
    \textbf{Unusual}\\ \textbf{Scene}
\end{minipage}
&
\begin{minipage}{.27\textwidth}
\centering
\includegraphics[height=30mm]{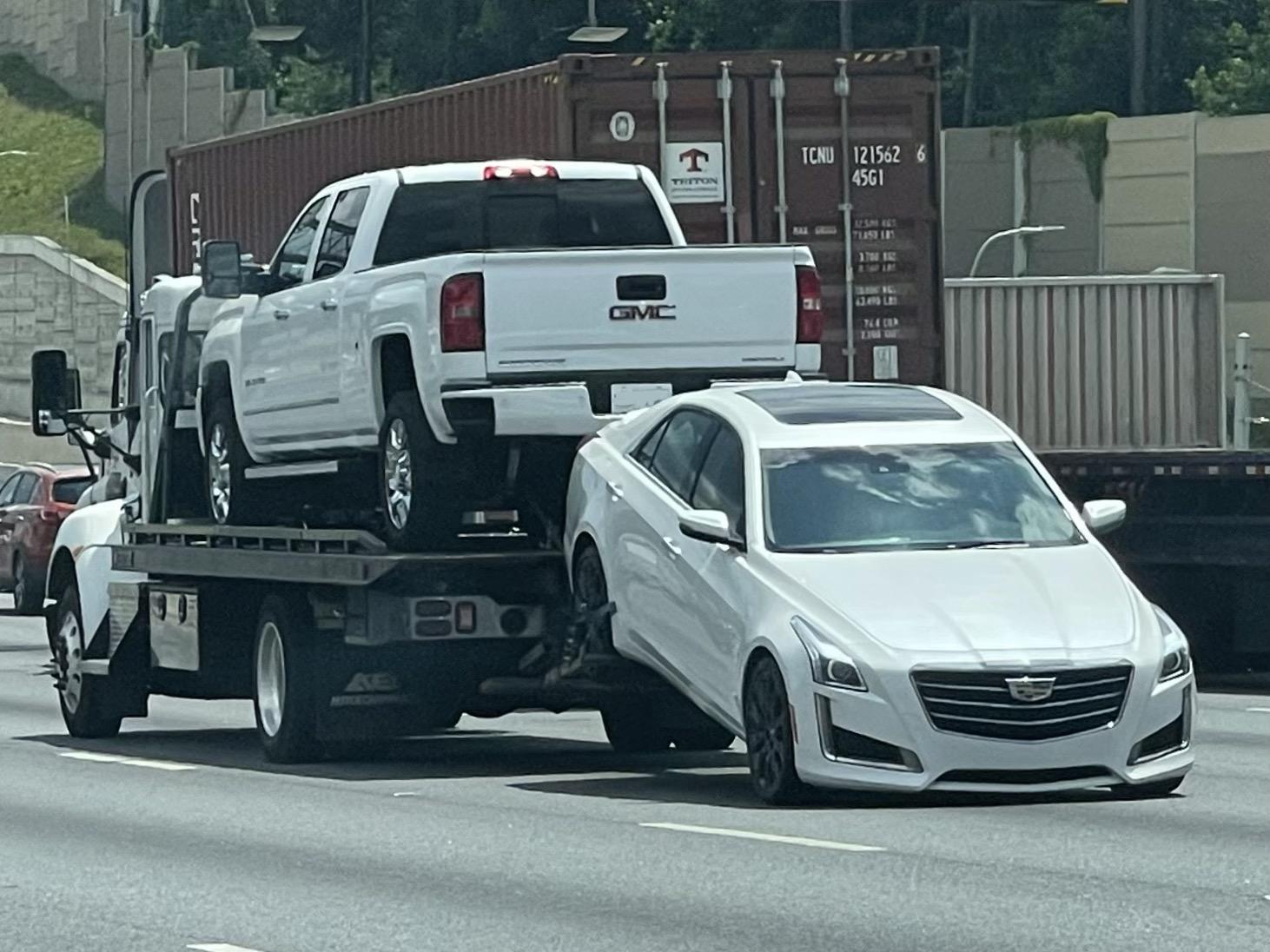}
\end{minipage}
& 
\begin{minipage}{.63\textwidth}
\textbf{GPT-4o-0806: }In this image, a flatbed tow truck is transporting a white GMC pickup truck. Unusually, a white Cadillac sedan is partially lodged underneath the rear of the pickup, \redmark{with its front end lifted off the ground}. The scene appears to be on a highway, with a shipping container visible in the background, indicating a busy roadway. The Cadillac's position suggests it may have collided with the pickup, resulting in this precarious arrangement. The surrounding environment includes concrete barriers and other vehicles, emphasizing the urban setting.\\
\color{red}{\textbf{Comment}: A misinterpretation of the scene, where it should be the rear end being lifted, not the front, as commonly assumed.}

\end{minipage}\\

\midrule

\begin{minipage}{.12\textwidth}
    \textbf{Important}\\ \textbf{Details}
\end{minipage}
&
\begin{minipage}{.25\textwidth}
\centering
\includegraphics[height=50mm]{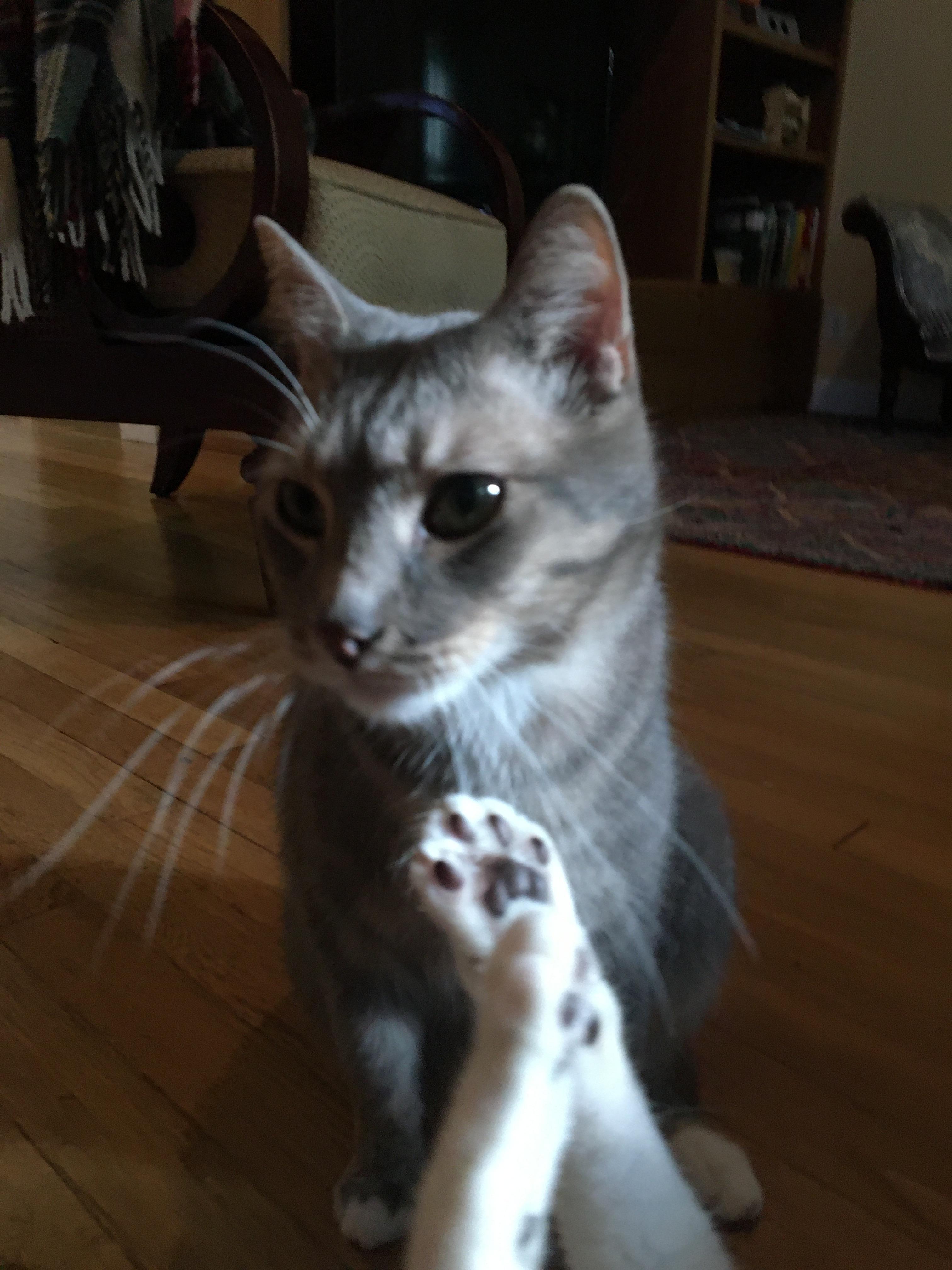}
\end{minipage}
& 
\begin{minipage}{.63\textwidth}
\textbf{Claude-3.5-Sonnet-0620: }This image shows a close-up of a gray cat sitting on a wooden floor. The cat has striking green eyes and is looking directly at the camera. Its fur appears to be a mix of light and dark gray, giving it a silvery appearance. The cat's whiskers are prominently visible. In the foreground, we can see \redmark{the cat's paw raised, showing its pink paw pads. This gives the impression} \redmark{that the cat might have been in the middle of grooming or was about to make a} \redmark{playful gesture}. The background of the image shows parts of what seems to be a living room. There's a glimpse of furniture, possibly a chair or sofa, and what looks like a bookshelf or cabinet in the distance. The room appears to be dimly lit, making the cat the main focus of the photograph. The wooden floor the cat is sitting on looks to be hardwood, and there's a hint of a rug or carpet visible at the edge of the frame. Overall, this image captures a candid and endearing moment of a domestic cat in its home environment, showcasing its curiosity and the close relationship between the photographer and the feline subject.\\
\color{red}{\textbf{Comment}: There should be two cats in the scene, and the paw in the foreground actually belongs to the other cat, not the one in the close-up.}

\end{minipage}\\
\midrule

\begin{minipage}{.12\textwidth}
    \textbf{Require}\\ \textbf{Association}
\end{minipage}
&
\begin{minipage}{.25\textwidth}
\centering
\includegraphics[height=31mm]{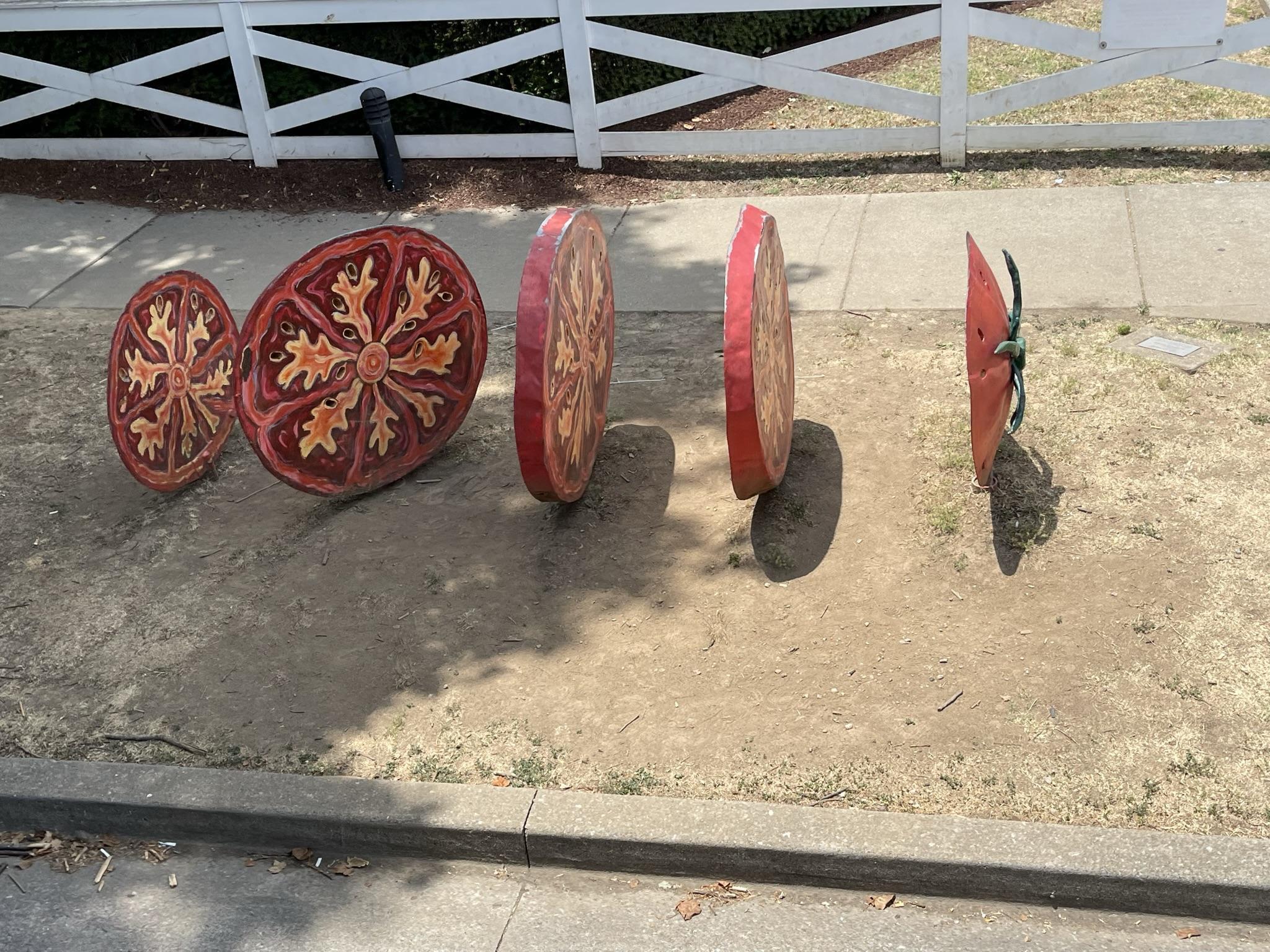}
\end{minipage}
& 
\begin{minipage}{.63\textwidth}
\textbf{GPT-4o-mini-0718: }A series of five artistic shields are displayed on a patch of dirt next to a sidewalk. \redmark{The shields, painted in vibrant reds and adorned with} \redmark{intricate designs that resemble leaves or vines, vary in size and orientation.} They are positioned in a line, suggesting movement, with the largest shield at the far left and gradually decreasing in size towards the right. A grassy area and a white wooden fence can be seen in the background, completing the scene with a touch of nature and structure. The setting appears to be an outdoor space, likely a park or community area, illuminated by bright sunlight.\\
\color{red}{\textbf{Comment}: It should be pointed out that this is a set of tomato slices, which is important for accurately describing the scene.}

\end{minipage}\\
\midrule

\begin{minipage}{.12\textwidth}
    \textbf{Identify}\\  \textbf{Clock Time}
\end{minipage}
&

\begin{minipage}{.25\textwidth}
\centering
\includegraphics[height=50mm]{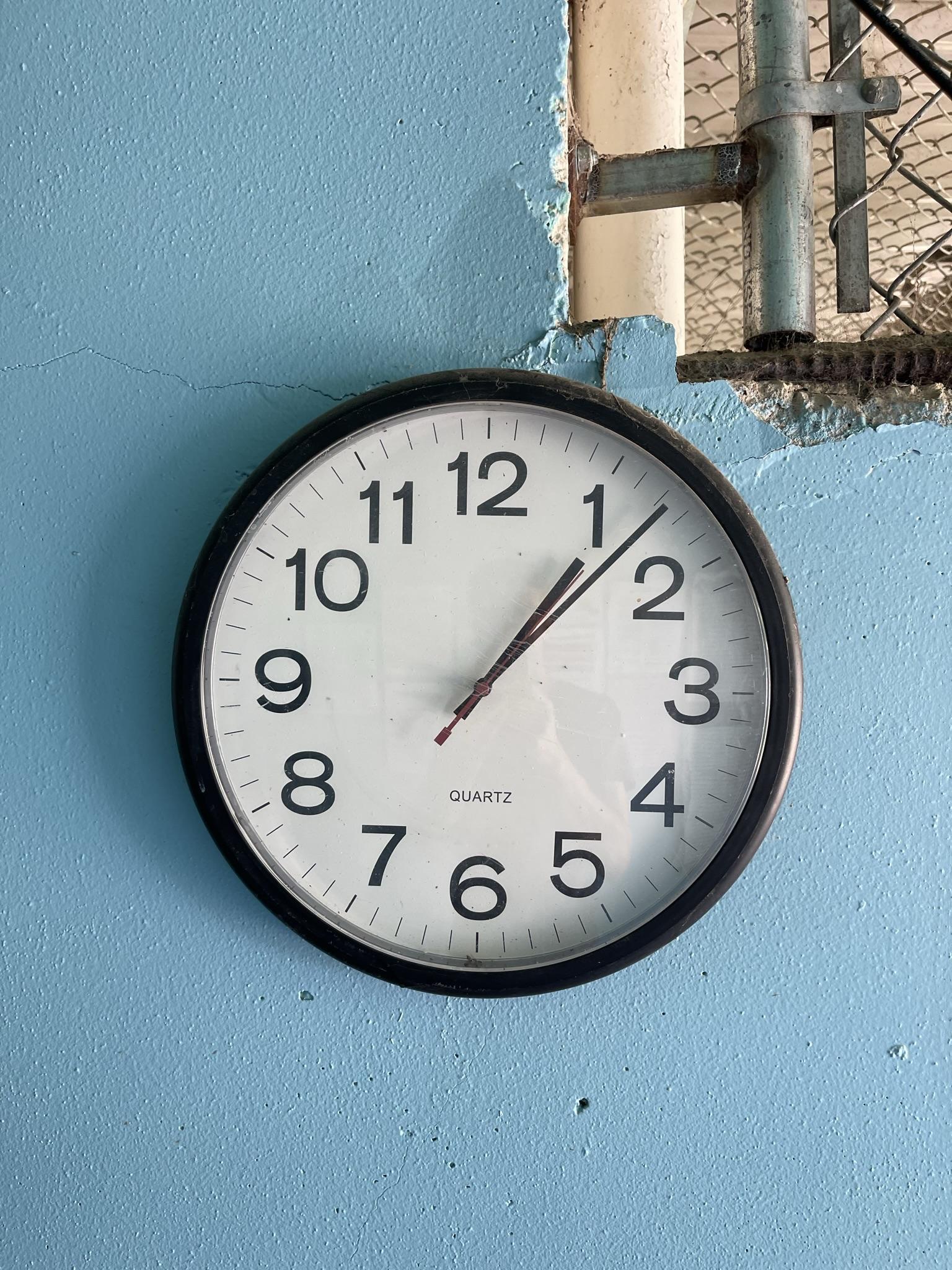}
\end{minipage}
& 
\begin{minipage}{.63\textwidth}
\textbf{Gemini-1.5-pro-002: }The image shows a close-up of an analog clock on a textured, light blue wall.  The clock has a black plastic or metal frame and a white face with black Arabic numerals. \redmark{The time displayed is approximately 1:53}.  The word "QUARTZ" is printed in small black letters below the center of the clock face. The wall shows some signs of wear and tear. There's a noticeable crack running diagonally across the top left portion of the frame, and some smaller imperfections and possible chipped areas scattered around the wall's surface. In the top right corner, partially obstructing the view of the wall, is the corner of a metal fence or gate. A galvanized metal pole is attached to the wall with a metal bracket. The wall appears damaged around the area where the pole is attached, with some of the surface material missing, revealing what looks like the underlying concrete or plaster.  A small section of chain-link fencing is also visible behind the pole.  This suggests the clock might be located in a more utilitarian or perhaps outdoor setting.\\
\color{red}{\textbf{Comment}: The correct time is about 1:07.}

\end{minipage}\\
\midrule

\begin{minipage}{.12\textwidth}
    \textbf{General}\\\textbf{Mistake}
\end{minipage}
&

\begin{minipage}{.25\textwidth}
\centering
\includegraphics[height=32mm]{latex/images/examples/test_02766.jpg}
\end{minipage}
& 
\begin{minipage}{.63\textwidth}
\textbf{LLaVA-1.6-34B: }The image captures a simple yet intriguing scene of \redmark{a stack of} \redmark{six rectangular concrete blocks arranged in a pyramid-like structure on a sandy} \redmark{ground}. The blocks, varying in shades of gray and brown, are stacked in such a way that \redmark{the smallest block is at the top, and the largest block is at the bottom}. The pyramid casts a shadow on the ground, adding depth to the scene. The sandy ground beneath the pyramid is scattered with small rocks and pebbles, providing a natural contrast to the man-made structure. The image is taken from a slightly elevated angle, giving a clear view of the pyramid and its surroundings. The overall composition of the image suggests a sense of balance and stability, as the pyramid stands firm on the sandy ground.\\
\color{red}{\textbf{Comment}: The description of the arrangement of the bricks in terms of quantity and size is incorrect. There are more than six equal-sized bricks.}

\end{minipage}\\
\bottomrule
\end{tabular}
}
\caption{Examples of failure cases.}.
\label{fig:failure-case}
\end{table*}

\section{Guidelines for Annotation}
\label{app:guideline}
Our guidelines for judging which of the two image captions is better are showed in Figure \ref{fig:guidelines}. 

While the guidelines provide a framework, they cannot account for all possible cases. Therefore, we encourage annotators to make informed judgments based on the specific circumstances and their own reasoning about which caption is more appropriate.

\section{Prompt Template}
\label{app:templete}
Building on \citet{chen2024mllm, li2024vlrewardbench}, we designed the VLM-as-a-Judge prompt for pairwise judgments, as shown in \Cref{fig:templete}.

\begin{figure*}[htbp]
\centering
\begin{tcolorbox}[colframe=black!50!gray, colback=white, coltitle=white, title=Guidelines, fonttitle=\bfseries\small, fontupper=\small 
]
As an annotator,your task is to judge which of the two image captions is better. Below are some guidelines for your reference:
    \begin{itemize}[leftmargin=8pt]
        \item \textbf{Precision:} The caption should accurately correspond to the content of the image, providing precise information about it. Common examples of imprecision include errors in color, quantity, spatial relationships, or the posture of people.
        \item \textbf{Informativeness:} Salient information in the image should be reflected in the caption. Since it is impossible to include every detail, you will need to subjectively judge which aspects of the image are important. For instance, describing an otter as "a small animal" is precise, but it is less informative than specifying "an otter".
        \item \textbf{Hallucination:} Captions that include descriptions of objects or elements that are clearly absent from the image should be significantly penalized.
        \item \textbf{Caption length:} Longer captions are not inherently better. For captions with equivalent informativeness, shorter ones are either better or at least not worse. Additionally, overly lengthy, verbose, or redundant expressions should be penalized.
        \item \textbf{Attention to detail:} Annotators should pay close attention to the details in the image to distinguish the quality of the descriptions.
        \item \textbf{Assistive description:} Imagine a visually impaired person asking you to describe the image for them. How would you convey the image to them?
        \item \textbf{Reverse thinking:} What image does the caption lead us to imagine? Does the caption effectively lead you to imagine the intended image?
        \item \textbf{Ties are acceptable:} If you find it genuinely difficult to determine which caption is better (e.g., both captions are excellent), marking a tie is acceptable.
    \end{itemize}
\end{tcolorbox}
\caption{Guidelines for Annotation} 
\label{fig:guidelines} 
\end{figure*}

\begin{figure*}[htbp]
\centering
\begin{tcolorbox}[colframe=black!50!gray, colback=white, coltitle=white, title=Template prompt of VLM-as-a-Judge, fonttitle=\bfseries\small, fontupper=\small]
 \textbf{(System Prompt)}

You are a highly capable multimodal AI assistant tasked with evaluating image captions.

 \textbf{(Instruction)}

Given an image and two candidate captions, you are require to determine which of the two captions is better.

\textbf{(Noticement)}

Below are some guidelines for your reference:
\begin{enumerate}[label=\arabic{enumi}.]
    \item **Precision**: The caption should accurately correspond to the content of the image, providing precise information about it. Common examples of imprecision include errors in color, quantity, spatial relationships, or the posture of people.
    
    \item **Informativeness**: Salient information in the image should be reflected in the caption. Since it is impossible to include every detail, you will need to subjectively judge which aspects of the image are important. For instance, describing an otter as ``a small animal'' is precise, but it is less informative than specifying ``an otter''.
    
    \item **Hallucination**: Captions that include descriptions of objects or elements that are clearly absent from the image should be significantly penalized.
    
    \item **Attention to detail**: Annotators should pay close attention to the details in the image to distinguish the quality of the descriptions.
    
    \item **Assistive description**: Imagine a visually impaired person asking you to describe the image for them. How would you convey the image to them?
    
    \item **Reverse thinking**: What image does the caption lead us to imagine? Does the caption effectively lead you to imagine the intended image?
    
    \item **Ties are acceptable**: If you find it genuinely difficult to determine which caption is better (e.g., both captions are excellent), marking a tie is acceptable.
\end{enumerate}
While the above guidelines provide a framework, they cannot cover all possible cases. Therefore, we encourage you to make **subjective judgments** based on the specific circumstances and your own reasoning about which caption is better.

\textbf{(Response Format)}

Format your response into two lines as shown below:
Reason: <your thoughts and reasoning process for the judgment>
Judgment: <Caption 1 is better>/<Caption 2 is better>/<Tie>

\end{tcolorbox}
\caption{Template prompt of VLM-as-a-Judge.} 
\label{fig:templete} 
\end{figure*}

\section{\caparenaA{} Leaderboard}
The current leaderboard of \caparenaA{} is provided in \Cref{tab:caparena-auto}.

\begin{table*}[ht]
\centering
\footnotesize
\renewcommand\arraystretch{1.1}
\resizebox{0.95\textwidth}{!}{%
\begin{tabular}{lcccccc}
\toprule
\textbf{Model} & \textbf{Score\_Avg$\uparrow$} & \textbf{Score\_GPT} & \textbf{Score\_Cog }& \textbf{Score\_CPM} & \textbf{Length\_Avg}\\
\midrule
\includegraphics[height=1em]{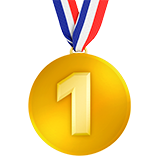}\includegraphics[height=1em]{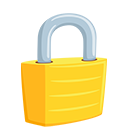}\textbf{\texttt{Gemini-1.5-pro-002}} & 56.17 & 29.0 & 61.0 &  78.5 & 168.6 \\
\includegraphics[height=1em]{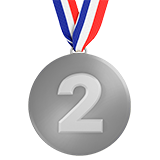}\includegraphics[height=1em]{latex/images/emoji/lock.png}\textbf{\texttt{GPT-4o-0806}} & 44.00 & 0 & 55.5 & 76.5 & 115.8 \\
\includegraphics[height=1em]{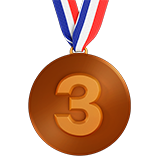}\includegraphics[height=1em]{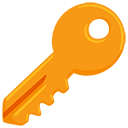}\textbf{\texttt{Qwen2.5-VL-72B-Instruct}} & 35.33 & -1.0 & 49.0 & 58.0 & 163.7 \\

\quad\includegraphics[height=1em]{latex/images/emoji/lock.png}\textbf{\texttt{Gemini-2.0-flash-exp}} & 30.83 & -2.0 & 39.5 & 55.0 & 417.0\\
\quad\includegraphics[height=1em]{latex/images/emoji/key.png}\textbf{\texttt{Ovis--2-34b}} & 27.00 & -15.0 & 33.5 & 62.5 & 120.2 \\
\quad\includegraphics[height=1em]{latex/images/emoji/lock.png}\textbf{\texttt{Claude-3.5-Sonnet-0620}} & 21.50 & -14.0 & 30.0 & 48.5 & 147.9 \\
\quad\includegraphics[height=1em]{latex/images/emoji/key.png}\textbf{\texttt{InternVL2-26B}}                 & 13.00 & -38.5 & 20.0 & 57.5 & 236.3 \\
\quad\includegraphics[height=1em]{latex/images/emoji/lock.png}\textbf{\texttt{GPT-4o-mini-0718}}             & 9.33 & -36.0 & 17.0 & 47.0 & 139.8 \\
\quad\includegraphics[height=1em]{latex/images/emoji/key.png}\textbf{\texttt{Ovis-1.6-27b}}                  & 3.00 & -49.5 & 14.5 & 44.0 & 94.2 \\
\quad\includegraphics[height=1em]{latex/images/emoji/lock.png}\textbf{\texttt{GLM-4V-Plus}}                  & -0.17 & -51.5 & 13.0 & 38.0 & 109.3 \\
\quad\includegraphics[height=1em]{latex/images/emoji/key.png}\textbf{\texttt{CogVLM2-llama3-chat-19B}}       & -8.50 & -56.5 & 0 & 31.0 & 115.9 \\
\quad\includegraphics[height=1em]{latex/images/emoji/key.png}\textbf{\texttt{Qwen2-VL-72B-Instruct}}         & -9.00 & -50.5 & -4.5 & 28.0 & 114.5 \\
\quad\includegraphics[height=1em]{latex/images/emoji/key.png}\textbf{\texttt{LLaVA-Onevision-72B-sft}}       & -12.33 & -57.5 & -6.0 & 26.5 & 200.9 \\
\quad\includegraphics[height=1em]{latex/images/emoji/key.png}\textbf{\texttt{LLama-3.2-vision-90B-Instruct}} & -25.67 & -72.0 & -13.0 & 8.0 & 160.3 \\
\quad\includegraphics[height=1em]{latex/images/emoji/lock.png}\textbf{\texttt{Hunyuan-standard-vision}}      & -26.00 & -63.0 & -19.0 & 4.0 & 354.1 \\
\quad\includegraphics[height=1em]{latex/images/emoji/key.png}\textbf{\texttt{InternVL2-5-8B}}                & -29.83 & -71.0 & -29.0 & 10.5 & 117.8 \\
\quad\includegraphics[height=1em]{latex/images/emoji/key.png}\textbf{\texttt{MiniCPM-V2.6-8B}}               & -38.00 & -80.0 & -34.0 & 0 & 106.7 \\
\quad\includegraphics[height=1em]{latex/images/emoji/key.png}\textbf{\texttt{Qwen2-VL-2B-Instruct}}          & -48.67 & -86.0 & -49.5 & -10.5 & 116.8 \\
\quad\includegraphics[height=1em]{latex/images/emoji/key.png}\textbf{\texttt{Qwen2-VL-7B-Instruct}}          & -49.00 & -78.0 & -59.0 & -10.0 & 97.8 \\
\quad\includegraphics[height=1em]{latex/images/emoji/key.png}\textbf{\texttt{LLaVA-1.6-34B}}                 & -67.50 & -92.0 & -53.5 & -57.0 & 124.8 \\
\quad\includegraphics[height=1em]{latex/images/emoji/key.png}\textbf{\texttt{cambrian-34b}}                  & -75.00 & -93.0 & -76.0 & -56.0 & 120.2 \\
\quad\includegraphics[height=1em]{latex/images/emoji/key.png}\textbf{\texttt{LLaVA-1.5-7B}}                  & -94.00 & -99.5 & -92.0 & -90.5 & 74.4 \\
\bottomrule
\end{tabular}
}
\caption{\caparenaA{} Leaderboard.}
\label{tab:caparena-auto}
\end{table*}

\end{document}